\documentclass[lettersize,journal]{IEEEtran}
\usepackage{amsmath,amsfonts}
\usepackage{array}
\usepackage[linesnumbered,boxed,ruled,commentsnumbered]{algorithm2e}
\usepackage{textcomp}
\usepackage{stfloats}
\usepackage{url}
\usepackage{verbatim}
\usepackage{graphicx}
\usepackage{cite}
\usepackage{booktabs}

\usepackage{amsmath,amsfonts}
\usepackage[psamsfonts]{amssymb}
\usepackage{amsxtra}
\usepackage{graphicx}
\usepackage{adjustbox}
\usepackage{multirow}
\usepackage{times}
\usepackage{helvet}
\usepackage{courier}
\usepackage[table]{xcolor} 
\usepackage{pifont} 
\usepackage{tikz}
\usetikzlibrary{positioning} 
\definecolor{lightpurple}{rgb}{0.87, 0.82, 1.0} 

\graphicspath{{./figure/} }
\DeclareGraphicsExtensions{.pdf,.jpeg,.png}

\newtheorem{myDef}{Definition}

\newcommand{\bluebold}[1]{\textbf{\textcolor{blue}{#1}}}
\newcommand{\redbold}[1]{\textbf{\textcolor{red}{#1}}}

\def\defeq{\stackrel{\triangle}{=}}

\makeatletter
\newcommand{\icircled}{\mathbin{\mathpalette\make@circled {\scriptsize{i}}}}
\newcommand{\make@circled}[2]{%
  \ooalign{$\m@th#1\smallbigcirc{#1}$\cr\hidewidth$\m@th#1#2$\hidewidth\cr}%
}
\newcommand{\smallbigcirc}[1]{%
  \vcenter{\hbox{\scalebox{1.05}{$\m@th#1\bigcirc$}}}%
}
\makeatother

\SetKwInput{KwInput}{Input}   
\SetKwInput{KwOutput}{Output} 
\SetKwRepeat{Do}{do}{while}   
\SetAlgoNlRelativeSize{0}     
\SetAlgoNlRelativeSize{-1}    

\begin{document}

\title{Joint Tensor and Inter-View Low-Rank Recovery for Incomplete Multiview Clustering}

\author{Jianyu Wang, Zhengqiao Zhao, Nicolas Dobigeon,~\IEEEmembership{Senior Member,~IEEE} and Jingdong Chen
\thanks{J. Wang, Z. Zhao and J. Chen are with Center of Intelligent Acoustics and
Immersive Communications, Northwestern Polytechnical University, China (email: alexwang96@mail.nwpu.edu.cn, zhengqiao.zhao@foxmail.com, jingdongchen@ieee.org).}
\thanks{N. Dobigeon is with University of Toulouse, IRIT/INP-ENSEEIHT,
CNRS, 2 rue Charles Camichel, BP 7122, 31071 Toulouse Cedex 7, France (e-mail: Nicolas.Dobigeon@enseeiht.fr).}
\thanks{Part of this work was supported by the Artificial Natural Intelligence Toulouse Institute (ANITI, ANR-19-PI3A-0004).}}

%

\maketitle

\begin{abstract}
Incomplete multiview clustering (IMVC) has gained significant attention for its effectiveness in handling missing sample challenges across various views in real-world multiview clustering applications. Most IMVC approaches tackle this problem by either learning consensus representations from available views or reconstructing missing samples using the underlying manifold structure. However, the reconstruction of learned similarity graph tensor in prior studies only exploits the low-tubal-rank information, neglecting the exploration of inter-view correlations. This paper propose a novel joint tensor and inter-view low-rank Recovery (JTIV-LRR), framing IMVC as a joint optimization problem that integrates incomplete similarity graph learning and tensor representation recovery. By leveraging both intra-view and inter-view low rank information, the method achieves robust estimation of the complete similarity graph tensor through sparse noise removal and low-tubal-rank constraints along different modes. Extensive experiments on both synthetic and real-world datasets demonstrate the superiority of the proposed approach, achieving significant improvements in clustering accuracy and robustness compared to state-of-the-art methods.
\end{abstract}


\section{Introduction}
\IEEEPARstart{M}{ultiview} data consists of samples captured from multiple perspectives or modalities \cite{bickel2004multi}, making it well-suited for classification and clustering analysis. It has important applications in fields such as image analysis \cite{kan2015multi}, video face recognition \cite{wu2013constrained}, and bioinformatics \cite{rappoport2018multi}. Compared to single-view approaches, which represent only one perspective and often provide a limited understanding of objects, multiview clustering (MVC) methods leverage complementary information from different views to obtain a more comprehensive and robust representation of the data \cite{chao2021survey}. By imposing key assumptions such as independence or correlation among different views, MVC has shown to enhance clustering performance by modeling deeper structures across views, overcoming the limitations of single-view methods, especially, the lack of holistic representations of the data \cite{fang2023comprehensive}.

MVC leverages feature representations from multiple perspectives or modalities, known as ``views'', to enhance clustering robustness surpassing the performance of single-view based approaches. Each view {$\mathbf{X}^{(v)}\in\mathbb{R}^{d_v\times n_v}$}, where {$n_v$} is the number of samples and $d_v$ is the dimension of $v$th view, captures distinct yet complementary characteristics of the data. Conventional MVC methods, such as co-training and co-regularization based approaches \cite{kumar2011co1, kumar2011co2}, seek to find a shared clustering structure by learning a consensus representation across different views. These approaches generally rely on the optimization problem
\begin{align}
 \min_{{\mathbf{T}},\{\mathbf{T}^{(v)},\mathbf{U}^{(v)}\}_{v=1}^V} & \sum_{v=1}^V \mathcal{F} \left( \mathbf{X}^{(v)} \Big| \mathbf{U}^{(v)} {\mathbf{T}^{(v)}} \right)  \label{eq:MVC} \\
 & + \alpha \psi\left( \mathbf{T}^{(v)}, {\mathbf{T}} \right) + \beta \mathcal{R}\left( \mathbf{U}^{(v)}; \mathbf{T}^{(v)} \right) \nonumber
\end{align}
where $\mathbf{T}^{(v)}\in\mathbb{R}^{K\times n_v}$ denotes the cluster
indicator matrix for the $v$th view, $K$ is the dimension of the subspace, ${\mathbf{T}}\in\mathbb{R}^{K\times n_v}$ is the shared cluster indicator matrix, and $\mathbf{U}^{(v)} \in \mathbb{R}^{d_v\times K}$ denotes a linear transformation matrix of the $v$th view. In \eqref{eq:MVC} the function $\mathcal{F}\left( \cdot | \cdot \right)$ measures the error between each view and the shared clustering structure, $\psi(\cdot)$ and $\mathcal{R}(\cdot)$ denote the regularization terms for enforcing consistency and preventing overfitting among views, respectively, and $\alpha$ and $\beta$ are parameters adjusting the amount of these regularizations.

The assumption of full data availability significantly limits the application
of MVC when some samples are missing in one or more views. In fact, in real-world applications, it is often difficult to obtain the complete data for all views of interest due to data collection limitations such as sensor failures, data corruption, or interrupted data acquisition processes. As a result, incomplete multiview clustering (IMVC) algorithms are drawing increasing attention \cite{trivedi2010multiview, wen2022survey}. In IMVC, representations from different views are often partially available, resulting in the loss of crucial information and difficulty in aligning views, significantly impacting clustering performance. The main challenge of IMVC lies in effectively utilizing the available data across all views while handling the missing samples \cite{xu2015multi, lin2021completer}. This includes effectively integrating incomplete cross-view information while preserving the underlying data structure and the relationships between samples \cite{zhan2018multiview}. To address this issue, two key strategies are commonly employed. The first is referred to as view completion, which aims to recover the missing samples by leveraging the available information from other views. For example, existing studies attempt to restore the completeness of the data by imputing or reconstructing the missing views \cite{liu2020efficient, yin2021incomplete, wen2018incomplete, zhang2018generalized, liu2024latent}, allowing conventional MVC techniques to be applied. However, imputation-based methods may introduce errors when the missing data is substantial or complex, leading to suboptimal clustering results. Alternatively, other methods bypass data imputation by directly learning a consensus latent representation from incomplete multiview data \cite{liu2022localized, deng2023projective,li2022refining, yang2024geometric}. By exploiting the intrinsic low-rank structure shared across views, prior studies have demonstrated their ability to effectively capture the underlying correlations between views and to recover the global structure of the data without a separate explicit data completion step \cite{wen2020generalized, wu2024low}. This not only mitigates the impact of missing views but also enhances the robustness of the clustering process by preserving important dependencies within the data.

Furthermore, many recent methods integrate both view completion and the low-rank structure across views to simultaneously recover missing samples and capture inter-view relationships \cite{lv2023joint, wang2021generative, hao2021multi}. For example, graph-based methods \cite{zhou2019consensus}, subspace learning \cite{yin2015incomplete, yin2017unified, liang2024robust}, anchor graph methods \cite{liu2022fast, li2023cross, yu2024dvsai}, and matrix factorization \cite{wen2023graph} are possible approaches to effectively exploit the shared and complementary information between views for IMVC \cite{liang2022incomplete, yin2023learning, yang2023cross}. Specifically, graph-based methods derive graph Laplacian matrices to encode the structural information of each view, which are subsequently combined into a shared graph Laplacian \cite{zhou2019consensus, li2022refining, yang2024geometric}. Even if such approaches successfully capture the relationships within each, they often struggle with cross-view consistency.
Subspace learning methods seek to learn a shared low-dimensional subspace that spans multiple views. While they can successfully capture the global structure, subspace learning based methods often suffer from computational complexity when dealing with high-dimensional data. Anchor graph methods improve the scalability by introducing a small number of anchor points, but this approximation may lead to loss of important fine-grained information. Matrix factorization techniques decompose a data matrix into low-rank factors, uncovering latent structures within the data. However, these methods often overlook higher-order correlations across multiple views. Despite this limitation, matrix factorization remains a widely used approach for learning latent representations that capture shared information across views \cite{wen2018incomplete, liu2022localized, deng2023projective, wen2023graph}.

Recent advances in tensor-based approacheshave demonstrated superior effectiveness in capturing both inter-view and intra-view information \cite{wen2021unified, zhang2023enhanced, hao2023tensor,wang2024incomplete}. Tensor formulations, particularly in graph and subspace methods, provide a powerful framework to model associations across views \cite{xing2024robust}, facilitating robust inter-view fusion. Unlike traditional graph-based methods that primarily capture low-rank structures within each view, tensor-based extensions account for low-rank information across views \cite{wan2024tensor, lv2023joint}, enabling a more comprehensive understanding of the complex relationships between views. This capability is crucial for robust clustering, as traditional methods often overlook these relationships by focusing solely on intra-view structures \cite{wang2024incomplete}.

Tensor-based methods are especially advantageous in incomplete multiview clustering scenarios. While existing approaches often treat each view independently \cite{liu2022fast}, tensor-based algorithms handle missing data by modeling complex inter-view interactions and maintaining scalability for high-dimensional data \cite{li2022high, shen2023robust}. Coupled tensor factorization \cite{9969457} captures shared and view-specific information, enabling robust clustering even in the presence of incomplete data. Tensor regularization techniques \cite{10004588} further improve generalizability by enforcing properties like sparsity and smoothness. Tensor decomposition methods, such as CP and Tucker decompositions \cite{kolda2009tensor}, address scalability challenges by reducing dimensionality and extracting essential features. Tensor completion techniques leverage low-rank structures to reconstruct missing entries, enhancing both data recovery and clustering performance \cite{zhang2023enhanced}. However, tensor completion based IMVC algorithms have rarely explored the low-rank information across multiple views. This work proposes to fill this gap by showing such information is essential to handle missing values. Although some prior studies have proposed to use low-tubal-rank constraints along different modes to capture such information \cite {shen2023robust}, their similarity graphs are predefined and highly depend on the estimation of the marginalized denoizer, thereby lacking of flexibility for real world heterogeneous data.

This paper proposes a novel joint tensor and inter-view low-rank recovery (JTIV-LRR) method for IMVC. The proposed approach represents multiview data as a tensor to capture high-order correlations among views. It recovers missing data and uncover the shared latent structure through low-tubal-rank constraints across different modes, which models different correlations along different modes. By jointly optimizing the tensor representation and inter-view low rank loss, the method learns robust multi-view fusion and improves the performance of downstream clustering analysis.

The major contributions of this work are summarized as follows.
\begin{itemize}
  \item We propose a new model, JTIV-LRR, which integrates tensor representation and low-rank recovery for IMVC. Our model can jointly capture the high-order correlations among different modes from similarity graph tensor, and recover missing views through inter-view low-rank structures.
  \item We designed multiple simulation experiments to demonstrate that as inter-view correlation increases, the reconstruction error across different modes with varying correlations decreases, revealing the underlying mechanism and applicability of the proposed methods.
  \item Extensive experiments on multiple benchmark datasets demonstrate that our method outperforms tensor based state-of-the-art approaches in terms of clustering accuracy and robustness to missing data.
\end{itemize}

\section{Notations and Preliminaries}

In the sequel of the paper, we use bold uppercase letters to denote matrices, e.g., $\mathbf{X}$, where its dimension can be denoted as
$\mathbf{X}\in\mathbb{R}^{I\times J}$, and the $(i,j)$-th element of $\mathbf{X}$ is denoted as $\left[ \mathbf{X} \right]_{i,j}$. Bold lowercase letters are used to denote the rows and columns of the matrices: $\mathbf{x}_{i}$ and $\mathbf{x}_{:,j}$ denote the $i$th row and $j$th column of the matrix $\mathbf{X}$, respectively. Similarly, we represent three-dimensional tensors using the calligraphic letters: the $(i,j,k)$th entry of the tensor $\mathcal{X}\in\mathbb{R}^{I\times J\times K}$ is denoted as $\left[ \mathcal{X} \right]_{i,j,k}$ and its $i$th horizontal, $j$th lateral, and $k$th frontal slices are denoted as $\mathcal{X}_{i,:,:}$, $\mathcal{X}_{:,j,:}$, and $\mathcal{X}_{:,:,k}$, respectively.
The discrete Fourier transformation (DFT) of three-dimension tensor $\mathcal{X}$ along the mode $3$ is written as $\overline{\mathcal{X}}= \mathsf{fft} \left( \mathcal{X},[], {3} \right)$. Similarly, the corresponding inverse DFT of $\overline{\mathcal{X}}$ can be computed as $\mathcal{X} = \mathsf{ifft} \left( \overline{\mathcal{X}},[], 3 \right)$

\begin{myDef}[Tensor Nuclear Norm \cite{semerci2014tensor}]
  The tensor nuclear norm $\|\mathcal{X}\|_{\circledast}$ is defined as follows:
\begin{align}
\|\mathcal{X}\|_{\circledast} \defeq \sum_{k=1}^K \left\| \overline{\mathcal{X}}_{:,:,k} \right\|_\ast = \sum_{k=1}^{K}\sum_{l=1}^{\min(I,J)} \sigma_l\left( \overline{\mathcal{X}}_{:,:,k} \right),
\end{align}
 where $\sigma_l$ denotes the $l$-th largest singular value.
\end{myDef}

\begin{myDef}[The block circulant operator \cite{kilmer2011factorization}]
Given a $3$-dimension tensor $\mathcal{X}\in\mathbb{R}^{I\times J\times K}$, the block circulant operator of $\mathcal{X}$, denoted as $\mathsf{bcirc}\left( \mathcal{X} \right)$, maps $\mathcal{X}$ to a block circulant matrix of size $IK\times JK$:
\begin{align}
 \mathsf{bcirc}(\mathcal{X}) \defeq \begin{bmatrix}
                        \mathcal{X}_{:,:,1}   & \mathcal{X}_{:,:,K} & \mathcal{X}_{:,:,K-1} & \cdots & \mathcal{X}_{:,:,2} \\
                        \mathcal{X}_{:,:,2}   & \mathcal{X}_{:,:,1} & \mathcal{X}_{:,:,K} & \cdots & \mathcal{X}_{:,:,3} \\
                        \vdots              & \vdots            & \vdots            & \ddots & \vdots \\
                        \mathcal{X}_{:,:,K}   & \mathcal{X}_{:,:,K-1} & \mathcal{X}_{:,:,K-2} & \cdots & \mathcal{X}_{:,:,1}
                      \end{bmatrix},
\end{align}
where each sub-block $\mathcal{X}_{:,:,k}$ is of size $I\times J$.

\end{myDef}

\begin{myDef}[Folding and unfolding operator \cite{kilmer2011factorization}]

The $\mathsf{unfold}$ operator converts $\mathcal{X}$ into a matrix with a size of $IK\times J$ and $\mathsf{fold}$ is its inverse operator:
\begin{align}
 \mathsf{unfold}\left( \mathcal{X} \right) \defeq \begin{bmatrix}
                                      \mathcal{X}_{:,:,1} \\
                                      \mathcal{X}_{:,:,2} \\
                                      \vdots \\
                                      \mathcal{X}_{:,:,K}
                                    \end{bmatrix}, \ \mathsf{fold}\left( \mathsf{unfold}\left( \mathcal{X} \right) \right) \defeq \mathcal{X}.
\end{align}

\end{myDef}

\begin{myDef}[t-Product\cite{kilmer2013third}]
Given two tensors, $\mathcal{X}\in\mathbb{R}^{I\times J\times K}$ and $\mathcal{Y}\in\mathbb{R}^{J\times L\times K}$, the tensor product (t-product) is defined as
\begin{align}
\mathcal{Z} \defeq \mathcal{X} \ast \mathcal{Y} = \mathsf{fold}\left( \mathsf{bcirc}\left( \mathcal{X} \right) \cdot \mathsf{unfold}\left( \mathcal{Y} \right) \right),
\end{align}
where $\mathcal{Z}\in\mathbb{R}^{I\times L\times K}$. This multiplication operation can also be performed efficiently using FFT \cite{8606166}
\begin{itemize}
  \item computing the FFT of both tensors $\mathcal{X}$ and $\mathcal{Y}$:
      \begin{align}
        \overline{\mathcal{X}} = \mathsf{fft}\left( \mathcal{X}, [], 3 \right), \overline{\mathcal{Y}} = \mathsf{fft}\left( \mathcal{Y}, [], 3 \right).
      \end{align}
  \item multiplying the matrices in the frequency domain:
      \begin{align}
       \overline{\mathcal{Z}}_{:,:,k} = \overline{\mathcal{X}}_{:,:,k} \cdot \overline{\mathcal{Y}}_{:,:,k}, \ k = 1,\cdots, K.
      \end{align}
  \item computing the inverse FFT of tensor $\overline{\mathcal{Z}}$
      \begin{align}
        \mathcal{Z} = \mathsf{ifft}\left( \overline{\mathcal{Z}}, [], 3 \right).
      \end{align}
\end{itemize}
\end{myDef}

\begin{myDef}[Tensor SVD (t-SVD) \cite{kilmer2011factorization}]
The tensor SVD of the $3$-dimension tensor $\mathcal{X}\in\mathbb{R}^{I\times J\times K}$ is denoted as the t-product of three sub-tensors based on the tensor singular value decomposition (t-SVD)
\begin{align}
\mathcal{X} \defeq \mathcal{U} \ast \mathcal{S} \ast \mathcal{V}^\mathsf{T},
\end{align}
where $\mathcal{U}\in\mathbb{R}^{I\times I\times K}$, and $\mathcal{V}\in\mathbb{R}^{J\times J\times K}$ are two orthogonal tensors and $\mathcal{S}\in\mathbb{R}^{I\times J\times K}$ is a f-diagonal tensor {\cite{kilmer2011factorization}}. $\mathcal{V}^\mathsf{T}$ is the transport of the tensor $\mathcal{V}$, defined by transposing each frontal slice and reversing the order of the slices.
\end{myDef}

\begin{myDef}[Tensor singular value thresholding (t-SVT) \cite{kilmer2011factorization, liu2012tensor, zhang2016exact}]
Let $\mathcal{X}\in\mathbb{R}^{I\times J\times K}$, and $\tau\geq 0$ be a thresholding parameter. The tensor singular value thresholding (t-SVT) operator, denoted by $\mathsf{D}_{\tau}\left(\mathcal{X}\right)$, is defined as
\begin{align}
\mathsf{D}_{\tau}\left( \mathcal{X} \right) \defeq \mathcal{U} \ast \mathcal{S}_\tau \ast \mathcal{V}^\mathsf{T},
\end{align}
where $\left[ \mathcal{S}_\tau \right]_{i,i,k} = \max\left( \left[ \mathcal{S} \right]_{i,i,k} - \tau, 0 \right)$ is a soft-thresholding to singular values of $\mathcal{S}$ with the threshold $\tau$.

\end{myDef}

\section{Method}

\begin{figure}
\centering
\begin{tikzpicture}
\node[inner sep=0pt]
    {\includegraphics[width=80mm]{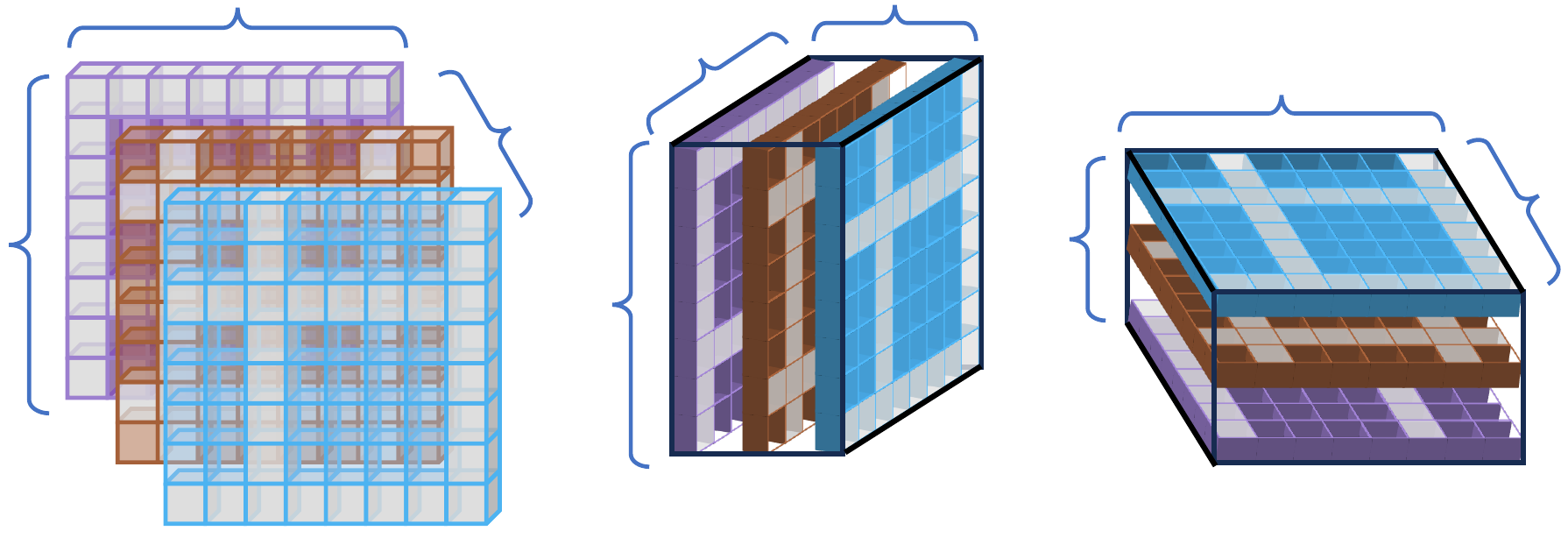}};
\draw  node[] at (-95pt,-45pt) {$\mathbf{G}^{(1)}$};
\draw  node[] at (-105pt,-35pt) {$\mathbf{G}^{(2)}$};
\draw  node[] at (-120pt,-20pt) {$\mathbf{G}^{(V)}$};
\draw  node[] at (-118pt,3pt) {$n$};
\draw  node[] at (-78pt,43pt) {$n$};
\draw  node[] at (-36pt,20pt) {$V$};
\draw  node[] at (-31pt,-5pt) {$n$};
\draw  node[] at (-20pt,30pt) {$n$};
\draw  node[] at (14pt,43pt) {$V$};
\draw  node[] at (12pt,-35pt) {$\mathbf{G}^{(1)}$};
\draw  node[] at (-5pt,-35pt) {$\mathbf{G}^{(2)}$};
\draw  node[] at (-22pt,-35pt) {$\mathbf{G}^{(V)}$};
\draw  node[] at (110pt,16pt) {$n$};
\draw  node[] at (70pt,34pt) {$n$};
\draw  node[] at (38pt,5pt) {$V$};
\draw  node[] at (115pt,-28pt) {$\mathbf{G}^{(V)}$};
\draw  node[] at (115pt,-13pt) {$\mathbf{G}^{(2)}$};
\draw  node[] at (115pt,2pt) {$\mathbf{G}^{(1)}$};
\end{tikzpicture}
\caption{Construction of similarity graph tensor $\mathcal{G}\in\mathbb{R}^{n\times n\times V}$ (left), and its two permutation format are $\mathcal{G}_{(1,3,2)} \in \mathbb{R}^{n\times V\times n}$ (middle) and $\mathcal{G}_{(3,2,1)}\in\mathbb{R}^{V\times n\times n}$. $\mathbf{G}^{(v)}\in\mathbb{R}^{n\times n}$ denotes the $v$th similarity graph matrix.
\label{fig1}}
\end{figure}
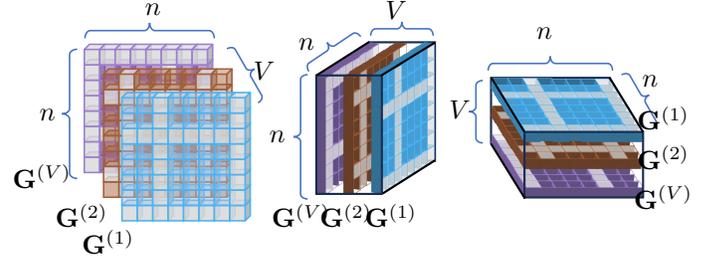

This section details the proposed JTIV-LRR method. Briefly, the incomplete data from each view are projected onto a latent subspace, enabling the minimization of reconstruction errors for missing views through models trained on the available multiview data. Subsequently, the latent subspaces are averaged to form a shared subspace, which facilitates the construction of multilevel self-representation graphs. These graphs effectively capture both consistent and view-specific information from the observed and reconstructed data. To further enhance correlation learning along different modes, a tensor nuclear norm regularizer is introduced to enforce low-rank structures both within each view and across different views. This is achieved by applying the nuclear norm regularization to three axis-permuted tensor graphs as depicted in Fig. \ref{fig1}, namely, mode 1 (horizontal slices), mode 2 (lateral slices), and mode 3 (frontal slices). Intuitively, horizontal and lateral slices represent inter-view correlations, while frontal slices reflect intra-view dependencies.

\subsection{Model formulation}

The method employed for recovering missing views involves constructing a unified latent representation derived from the observable multiview data. The approach operates under the assumption that all views of a given instance can be generated from a shared latent structure, ensuring consistency among different views. The ultimate goal is to utilize these insights to reconstruct the missing views effectively.
{First,
the complete sample matrix denoted $\tilde{\mathbf{X}}^{(v)} \in \mathbb{R}^{d_v \times n}$ is related to each corresponding observed view $\mathbf{X}^{(v)} \in \mathbb{R}^{d_v \times n_v}$ through an alignment matrix $\mathbf{W}^{(v)} \in \mathbb{R}^{n \times n_v}$
\begin{align}\label{subspace1}
\tilde{\mathbf{X}}^{(v)} \defeq \mathbf{X}^{(v)} \mathbf{W}^{(v)},
\end{align}
where $\mathbf{W}^{(v)}$ is the semi-orthogonal matrix defined as
\begin{align}
\mathbf{W}^{(v)}_{ij} \defeq
\begin{cases}
  1, & \text{if the } i\text{th sample in } \tilde{\mathbf{X}}^{(v)} \text{ corresponds to the} \nonumber\\
  & j\text{th sample in } \mathbf{X}^{(v)}, \nonumber\\
0, & \text{otherwise}
\end{cases}
\end{align}
with $\mathbf{W}^{(v)\mathsf{T}} \mathbf{W}^{(v)} = \mathbf{I}_n$. The matrix $\mathbf{W}^{(v)}$ ensures that $\tilde{\mathbf{X}}^{(v)}$ is aligned with the complete samples across all views by filling missing samples with zeros. Thus, the elements of $\tilde{\mathbf{X}}^{(v)}$ do not directly represent the missing values but rather indicate the positions within the complete view where the missing elements are located, as specified by the zero entries in $\mathbf{W}^{(v)}$. The exact values of these missing elements are unknown, and their positions are known: the missing entries in $\tilde{\mathbf{X}}^{(v)}$ correspond to the positions where $\mathbf{W}^{(v)}$ has zero values. Besides
the complete view matrix $\tilde{\mathbf{X}}^{(v)} $ follows a self-representation model, similar to an archetypal analysis \cite{cutler1994archetypal}
\begin{align} \label{eq:AA}
\tilde{\mathbf{X}}^{(v)} \defeq \tilde{\mathbf{X}}^{(v)} \tilde{\mathbf{G}}^{(v)}
\end{align}
where ${\tilde{\mathbf{G}}^{(v)}} \in \mathbb{R}^{n \times n}$ is the view-specific representation matrix, referred to as graph in what follows. For each view, the  similarity graph $ \tilde{\mathbf{G}}^{(v)}$ only captures relationships among the observed samples, making it insufficient for accurate modeling of the entire sample set. However, combining   \eqref{subspace1} and \eqref{eq:AA} leads to
\begin{align}
    \mathbf{X}^{(v)} = \mathbf{X}^{(v)} \mathbf{W}^{(v)} \tilde{\mathbf{G}}^{(v)} {\mathbf{W}^{(v)\mathsf{T}}}.
\end{align}
The resulting complete similarity graphs $\{ \tilde{\mathbf{G}}^{(v)} \}_{v=1}^V $ are then stacked along the third dimension to form the graph tensor $\tilde{\mathcal{G}} = \tilde{\mathcal{G}}\left( \tilde{\mathbf{G}}^{(1)}, \dots, \tilde{\mathbf{G}}^{(V)} \right) \in \mathbb{R}^{n \times n \times V}$ that fully encodes the latent relationship between the observed multiview data. This graph tensor is assumed to follow a robust low-rank decomposition
\begin{align}
\tilde{\mathcal{G}} = \tilde{\mathcal{L}} + \tilde{\mathcal{S}},
\end{align}
where $\tilde{\mathcal{L}} = \tilde{\mathcal{L}} \left( \tilde{\mathbf{L}}^{(1)}, \dots, \tilde{\mathbf{L}}^{(V)} \right) \in \mathbb{R}^{n\times n\times V}$ and $\tilde{\mathcal{S}}\in\mathbb{R}^{n\times n\times V} =  \tilde{\mathcal{S}} \left( \tilde{\mathbf{S}}^{(1)}, \dots, \tilde{\mathbf{S}}^{(V)} \right)$ denote the low-rank and sparse tensors, respectively. The conventional way of promoting the low-rankness of $\tilde{\mathcal{L}}$ consists in penalizing the tensor nuclear norm along the third view of the $n\times n\times V$ graph tensor $\tilde{\mathcal{L}}$. This approach is known to effectively capture intra-view data structures but may overlook the sample-specific patterns across views. Building upon the approach in \cite{shen2023robust}, this work advances further by simultaneously promoting the low-rankness of the reordered counterparts of this tensor, enabling direct modeling of interactions between samples and views. By incorporating low-rank structures across multiple dimensions, this strategy completes missing information through not only  inter-view but also intra-view information. It establishes a robust representation of the multiview data, thereby improving expected performance of the downstream clustering tasks. More precisely, the proposed JTIV-LRR method consists in first solving the following optimization problem
\begin{align}\label{model1}
    & \min_{\boldsymbol{\Theta}}  \left\| \tilde{\mathcal{S}} \right\|_1 + \lambda_1\left\| \tilde{\mathcal{L}} \right\|_\circledast + {\lambda_2}\left\| \tilde{\mathcal{L}}_{[2]} \right\|_\circledast + {\lambda_3} \left\| \tilde{\mathcal{L}}_{[3]} \right\|_\circledast  \nonumber \\
    & \mbox{s.t.} \ \ \tilde{\mathcal{G}} = \tilde{\mathcal{L}} + \tilde{\mathcal{S}} \ \ \text{and}\ \ {\mathbf{X}}^{(v)} = {\mathbf{X}}^{(v)} {\mathbf{W}^{(v)}} \tilde{\mathbf{G}}^{(v)} {\mathbf{W}^{(v)\mathsf{T}}}
\end{align}
where ${\boldsymbol{\Theta}} = \{ \tilde{\mathcal{G}}, \tilde{\mathcal{L}}, \tilde{\mathcal{S}}  \}$ contains the unknown tensors to be estimated, and $\lambda_1, {\lambda_2}$ and ${\lambda_3}$ are hyperparameters controlling the balance among the different terms. In \eqref{model1}, low-rank penalizations are applied to mode-permuted couterpart of the initial tensors adopting the convenient notations
\begin{align}
    \tilde{\mathcal{L}}_{[1]}& = \tilde{\mathcal{L}}\\
    \tilde{\mathcal{L}}_{[2]} &= \mathsf{permute}\left(\tilde{\mathcal{L}},[1,3,2]\right) = \tilde{\mathcal{L}}_{(1,3,2)} \\
        \tilde{\mathcal{L}}_{[3]} &= \mathsf{permute}\left(\tilde{\mathcal{L}},[3,2,1]\right)= \tilde{\mathcal{L}}_{(3,2,1)}.
\end{align}
Then, once the low-rank tensor $\tilde{\mathcal{L}} $ has been identified, the consistency matrix $\mathbf{C}\in\mathbb{R}^{n\times n}$ required for the downstream clustering analysis can be obtained as
\begin{align}
\mathbf{C} = \frac{1}{V} \sum_{v=1}^V \frac{\left| \tilde{\mathbf{L}}^{(v)} \right|+ \left| {\tilde{\mathbf{L}}^{(v) \mathsf{T}}} \right| }{2}
\end{align}
which represents the average of the symmetrized absolute values of each view low-rank representations $\tilde{\mathbf{L}}^{(v)}$. It ensures that
$\mathbf{C}$ effectively captures the shared information across all views, providing a unified metric for evaluating the low-rank structure of multiview data.  The cluster matrix can estimated using any off-the-shelf clustering method such as $K$-means. The next section described the algorithmic strategy to solve the problem \eqref{model1}.}

\subsection{Optimization}

Based on the alternating direction minimization method of multipliers (ADMM), auxiliary variables $\mathcal{Z}_1$, $\mathcal{Z}_2$, and $\mathcal{Z}_3$ are introduced and the problem \eqref{model1} is rewritten as
\begin{align}
     \min_{\boldsymbol{\Theta}} & \left\| \tilde{\mathcal{S}} \right\|_1 + {\lambda_1}\left\| \mathcal{Z}_1 \right\|_\circledast  + {\lambda_2} \left\| \mathcal{Z}_2 \right\|_\circledast + {\lambda_3} \left\| \mathcal{Z}_3 \right\|_\circledast  \\
     \mbox{s.t.} \ \ &  \tilde{\mathcal{G}}=\tilde{\mathcal{L}} + \tilde{\mathcal{S}}, \ {\mathbf{X}}^{(v)} = {\mathbf{X}}^{(v)} {\mathbf{W}^{(v)}} \tilde{\mathbf{G}}^{(v)} {\mathbf{W}^{(v)\mathsf{T}}}, \nonumber \\
    &  \mathcal{Z}_1 = \tilde{\mathcal{L}}, \ \mathcal{Z}_2 = \tilde{\mathcal{L}}_{[2]} \ \text{and} \  \mathcal{Z}_3 = \tilde{\mathcal{L}}_{[3]}.\nonumber
\end{align}
The augmented Lagrangian function of the above problem is\begin{align}
    f\left( \boldsymbol{\Theta} \right) = &  \left\| \mathcal{S} \right\|_1 + {\lambda_1} \left\| \mathcal{Z}_1 \right\|_\circledast + {\lambda_2} \left\| \mathcal{Z}_2 \right\|_\circledast + {\lambda_3} \left\| \mathcal{Z}_3 \right\|_\circledast \nonumber \\
    &  + \sum_v \left\langle \mathbf{F}^{(v)}, \mathbf{X}^{(v)} - {\mathbf{X}}^{(v)} {\mathbf{W}^{(v)}} \tilde{\mathbf{G}}^{(v)} {\mathbf{W}^{(v)\mathsf{T}}} \right\rangle \nonumber \\
    & + \left\langle \mathcal{J}_1, \tilde{\mathcal{G}} - \tilde{\mathcal{L}} - \tilde{\mathcal{S}} \right\rangle + \left\langle \mathcal{J}_2, \tilde{\mathcal{L}} - \mathcal{Z}_1 \right\rangle \nonumber \\
    & + \left\langle \mathcal{J}_3, \tilde{\mathcal{L}}_{[2]} - \mathcal{Z}_2 \right\rangle + \left\langle \mathcal{J}_4, \tilde{\mathcal{L}}_{[3]} - \mathcal{Z}_3 \right\rangle  \nonumber \\
    & + \frac{\rho}{2}\left\| \mathbf{X}^{(v)} - \mathbf{X}^{(v)} {\mathbf{W}^{(v)}} \tilde{\mathbf{G}}^{(v)} {\mathbf{W}^{(v)\mathsf{T}}} \right\|_\mathsf{F}^2  \nonumber \\
    & + \frac{\rho}{2} \left\| \tilde{\mathcal{G}} - \tilde{\mathcal{L}} - \tilde{\mathcal{S}} \right\|_\mathsf{F}^2 + \frac{\rho}{2} \left\| \tilde{\mathcal{L}} - \mathcal{Z}_1 \right\|_\mathsf{F}^2  \nonumber \\
    & + \frac{\rho}{2} \left\| \tilde{\mathcal{L}}_{[2]} - \mathcal{Z}_2 \right\|_\mathsf{F}^2 + \frac{\rho}{2} \left\| \tilde{\mathcal{L}}_{[3]} - \mathcal{Z}_3 \right\|_\mathsf{F}^2
\end{align}
where $\{\mathbf{F}^{(v)}\}_{v=1}^V$, $\mathcal{J}_1$, $\mathcal{J}_2$, $\mathcal{J}_3$, and $\mathcal{J}_4$ are Lagrangian multipliers, and $\rho\geq 0$ is a penalty factor. $\left\langle \cdot, \cdot \right\rangle$ denotes the sum of the element-wise products.
Consequently, the optimization process can be separated into the following subproblems. Note that,  to lighten the notations, the algorithm iteration index $k$ is omitted when possible without ambiguity.

\subsubsection{$\tilde{\mathbf{G}}^{(v)}$-Subproblem}
By introducing the  variables
\begin{align}
 & \mathbf{Q}^{(v)} = \mathbf{X}^{(v)} + \frac{\mathbf{F}^{(v)}}{\rho}\\
 & \mathbf{P}^{(v)} = \tilde{\mathbf{L}}^{(v)} + \tilde{\mathbf{S}}^{(v)} - \frac{{\mathbf{J}_1}^{(v)}}{\rho}
\end{align}
the optimization subproblem associated with $\tilde{\mathbf{G}}^{(v)}$ is
\begin{align}
  & \min_{\tilde{\mathbf{G}}^{(v)}} \left\| \mathbf{Q}^{(v)} - \mathbf{X}^{(v)}{\mathbf{W}^{(v)}}\tilde{\mathbf{G}}^{(v)} {\mathbf{W}^{(v)\mathsf{T}}} \right\|_\mathsf{F}^2  + \left\| \tilde{\mathbf{G}}^{(v)} - \mathbf{P}^{(v)} \right\|_\mathsf{F}^2.
\end{align}
The updating rule thus consists in solving  the following Sylvester equation \cite{10.1145/361573.361582,7163298}
\begin{align}\label{eq:sylvester}
     &{\mathbf{W}^{(v)\mathsf{T}}} {\mathbf{X}^{(v)\mathsf{T}}} \mathbf{X}^{(v)} {\mathbf{W}^{(v)}} \tilde{\mathbf{G}}^{(v)} + \tilde{\mathbf{G}}^{(v)}\mathbf{E}^{(v)}  = \mathbf{M}^{(v)}
\end{align}
Where
\begin{align}
    \mathbf{M}^{(v)}  &= \left( {\mathbf{W}^{(v)\mathsf{T}}} {\mathbf{X}^{(v)\mathsf{T}}} \mathbf{Q}^{(v)} {\mathbf{W}^{(v)}} + \mathbf{P}^{(v)} \right) \mathbf{E}^{(v)}  \nonumber \\
    \mathbf{E}^{(v)}                &= \left( {\mathbf{W}^{(v)\mathsf{T}}} {\mathbf{W}^{(v)}} \right)^{-1}.
\end{align}
The solution to this Sylvester equation can be computed by any off-the-shelf dedicated solver. It is clear that the reconstructed low-rank structure $\tilde{\mathbf{G}}^{(v)}$ remains consistent with the constraints imposed by the original data and the low-rank representation learned in the previous iterations.

\subsubsection{$\mathcal{Z}_m$-Subproblems}
The optimization subproblems for $\mathcal{Z}_m$ ($m=1,\ldots,3$) corresponds to different mode rearrangements of the low-rank tensor. Their updating rules only differ by the mode permutation imposed to the low-rank components and can be implemented separately. A generic common formulation of the subproblems writes
\begin{align}
    \min \lambda_m \left\| \mathcal{Z}_m \right\|_{\circledast} +  \frac{\rho}{2} \left\|  \tilde{\mathcal{L}}_{[m]} -  \mathcal{Z}_m+  \frac{\mathcal{J}_{m+1}}{\rho} \right\|_\mathsf{F}^2.
\end{align}
The solution can be obtained using the singular value thresholding (SVT) operator \cite{doi:10.1137/080738970, 8606166}
\begin{align}
    \overline{\mathcal{Z}_m} = \mathsf{D}_{\frac{\lambda_m}{\rho}} \left( \overline{\tilde{\mathcal{L}}_{[m]}} + \frac{\overline{\mathcal{J}_2}}{\rho} \right).
\end{align}
The SVT shrinks the singular values of the input matrix by a threshold
$\tau$, promoting low-rankness in the updated variables. These updates ensure that the low-rank representation captures information from all three modes of the tensor, thereby improving the quality of the final reconstruction.

\subsubsection{$\tilde{\mathcal{L}}$-Subproblem}
The update of the low-rank component $\tilde{\mathcal{L}}$ is obtained by solving the following minimization problem:
\begin{align}
    & \min_{\tilde{\mathcal{L}}} \left\|  \tilde{\mathcal{L}} - \mathcal{Z}_1  +  \frac{\mathcal{J}_2}{\rho} \right\|_\mathsf{F}^2 + \left\|  \tilde{\mathcal{L}}_{[2]}  -  \mathcal{Z}_2  +  \frac{\mathcal{J}_3}{\rho} \right\|_\mathsf{F}^2 \nonumber \\
    & + \left\| \tilde{\mathcal{L}}_{[3]}  -  \mathcal{Z}_3  +  \frac{\mathcal{J}_4}{\rho} \right\|_\mathsf{F}^2 +  \left\| \tilde{\mathcal{G}} - \tilde{\mathcal{L}} - \tilde{\mathcal{S}} + \frac{\mathcal{J}_1}{\rho} \right\|_\mathsf{F}^2
\end{align}
where each term corresponds to a different constraint in the optimization problem. This quadratic problem can be solved explicitly and the updating rule is
\begin{align}
    \tilde{\mathcal{L}} = & \frac{1}{4} \left( \tilde{\mathcal{G}} - \tilde{\mathcal{S}} + \mathcal{Z}_1 +  \mathcal{Z}_2+  \mathcal{Z}_3 + \frac{\mathcal{J}_1 - \mathcal{J}_2 - \mathcal{J}_3 - \mathcal{J}_4}{\rho} \right).
\end{align}
This solution effectively averages the contributions from each low-rank variable and the Lagrange multipliers, ensuring that the updated low-rank component $\tilde{\mathcal{L}}$ is consistent with all constraints and minimization goals.

\begin{algorithm}[t]
\caption{JTIV-LRR}\label{Alg1}
\KwInput{Data matrix $\{\mathbf{X}^{(v)}\}_{v=1}^V$, index matrix $\mathbf{W}^{(v)}$, parameters  ${\lambda_1}$, ${\lambda_3}$, and ${\lambda_2}$.}
\KwOutput{The consistency matrix $\mathbf{C}$.}
\SetAlgoNlRelativeSize{0}     
\SetKwBlock{Begin}{Initialize}{}
\Begin{
  Construct $\mathbf{W}^{(v)}$ as (16); \\
  Initialize $\mathcal{J}_1 = 0, \mathcal{J}_2 = 0, \mathcal{J}_3 = 0, \mathcal{J}_4 = 0, \mathbf{F}^{(v)} = 0, \rho = 10^{-4}$;
}

\While{not converge}{
  Update $\tilde{\mathbf{G}}^{(v)}$ by Eq. (24); \\
  Update $\mathcal{Z}_1,\mathcal{Z}_2,\mathcal{Z}_3$ by Eq. (28); \\
  Update $\tilde{\mathcal{L}}$ by Eq. (30); \\
  Update $\tilde{\mathcal{S}}$ by Eq. (32); \\
  Update $\mathbf{F}^{(v)}, \mathcal{J}_1, \mathcal{J}_2, \mathcal{J}_3, \mathcal{J}_4$ by Eq. (34), (35), (36), (37), (38);
}
\KwRet{Consistency matrix $\mathbf{C}$.}
\end{algorithm}

\subsubsection{$\tilde{\mathcal{S}}$-Subproblem}
The subproblem for updating the sparse component $\tilde{\mathcal{S}}$ is
\begin{align}\label{sparseSub}
    &\min_{\tilde{\mathcal{S}}} \frac{1}{\rho} \left\| \tilde{\mathcal{S}} \right\|_1  +  \frac{1}{2} \left\| \tilde{\mathcal{G}}  -  \tilde{\mathcal{L}}  -  \tilde{\mathcal{S}}  +  \frac{\mathcal{J}_1}{\rho} \right\|_\mathsf{F}^2.
\end{align}
This subproblem can be solved using a soft-thresholding \cite{10.1111/j.2517-6161.1996.tb02080.x}
\begin{align}
 \tilde{\mathcal{S}} = \mathsf{ST}_{\frac{1}{\rho}} \left( \tilde{\mathcal{G}} - \tilde{\mathcal{L}} + \frac{\mathcal{J}_1}{\rho} \right)
\end{align}
where $\mathsf{ST}_\tau\left(\cdot\right)$ is the element-wise shrinkage operator
\begin{align}
 \mathsf{ST}_\tau\left( x \right) = \mathsf{sign}(x)\cdot \max(|x|-\tau, 0).
\end{align}

\subsubsection{Update the Lagrange multipliers }
The Lagrange multipliers $\mathbf{F}^{(v)}$, $\mathcal{J}_1$, $\mathcal{J}_2$, $\mathcal{J}_3$ and $\mathcal{J}_4$ are updated iteratively to ensure that the constraints imposed in the optimization problem are satisfied. The updating rules are given by
\begin{align}
    \mathbf{F}^{(v)}_{k+1} &=  \mathbf{F}^{(v)}_k \\
                            & \ \ \ + \rho  \left( \mathbf{X}^{(v)} - \mathbf{X}^{(v)} {\mathbf{W}^{(v)}} \tilde{\mathbf{G}}^{(v)}_{k+1} {\mathbf{W}^{(v)\mathsf{T}}} \right) \nonumber \\
    \mathcal{J}_{1,k+1} &= \mathcal{J}_{1,k} + \rho\left( \tilde{\mathcal{G}}_{k+1} - \tilde{\mathcal{L}}_{[1],k+1} - \tilde{\mathcal{S}}_{k+1} \right) \\
    \mathcal{J}_{2,k+1} &= \mathcal{J}_{2,k} + \rho\left( \tilde{\mathcal{L}}_{[1],k+1} - \mathcal{Z}_{1,k+1} \right) \\
    \mathcal{J}_{3,k+1} &= \mathcal{J}_{3,k} + \rho\left( \tilde{\mathcal{L}}_{[2],k+1} - \mathcal{Z}_{2,k+1} \right) \\
     \mathcal{J}_{4,k+1} &= \mathcal{J}_{4,k} + \rho\left( \tilde{\mathcal{L}}_{[3],k+1} - \mathcal{Z}_{3,k+1} \right).
\end{align}

\subsection{Computational Complexity}
Hereafter, we analyze the computational complexity of the proposed algorithmic procedure JTIV-LRR sketched in Algo. \ref{Alg1}. At each iteration, solving the Sylvester equation and performing the matrix inverse operation to update the similarity graph $\tilde{\mathbf{G}}^{(v)}$ requires $\mathcal{O}(n^3)$ operations. The index matrix $\mathbf{W}^{(v)}$ is fixed, and the computation of $\mathbf{W}^{(v)^\mathsf{T}} \mathbf{W}^{(v)}$ is performed outside the loop. For updating $\mathcal{Z}$, $\mathcal{Z}_2$ and $\mathcal{Z}_3$, the computational complexity is dominated by the FFT, inverse FFT, and SVD operations. The FFT and inverse FFT operations have a complexity of $\mathcal{O}(Vn^2\log(n))$, while the SVD operation requires $\mathcal{O}(V^2n^2)$.
Since the other steps only involve basic matrix operations with lower computational costs than the aforementioned calculations, their complexity is negligible. Therefore, the computational complexity of one iteration of the algorithm is approximately given by $\mathcal{O} \big(n^3 + Vn^2\log(n) + V^2n^2\big)$.

\section{Experiments}

This section  provides an in-depth explanation of the  experiments conducted to assess the effectiveness of the proposed JTIV-LRR method. Furthermore, this comprehensive performance analysis is completed with  an ablation study to evaluate the impact of the model parameter and with an empirical convergence analysis.

\begin{table}[!t]
\centering
\caption{Datasets in Experiments}
\begin{tabular}{lcccc} 
\toprule
\textbf{Datasets} & \textbf{Size} & \textbf{Classes} & \textbf{Views} & \textbf{Dimensionality} \\
\midrule
3Sources     & 169  & 1   & 3  & 3560, 3631, 3068 \\
100Leaves    & 1600 & 100 & 3  & 64, 64, 64 \\
ORL          & 400  & 40  & 3  & 4096, 3304, 6750 \\
ProteinFold  & 694  & 27  & 12 & 27, $\cdots$, 27 \\
Scene        & 4485 & 15  & 3  & 20,   59,   40   \\
Mfeat        & 2000 & 10  & 6  & 216, 76, 64, 6, 240, 47 \\
Caltech101-7 & 1474 & 7   & 6  & 48, 40, 254, 1984, 512, 928 \\
\bottomrule
\end{tabular}
\label{tab:datasets}
\end{table}

\subsection{On the low-rankness of mode-permuted tensors}

\begin{figure}[!t]
    \centering
    \begin{minipage}[b]{0.24\textwidth} 
        \centering
        \includegraphics[width=\textwidth]{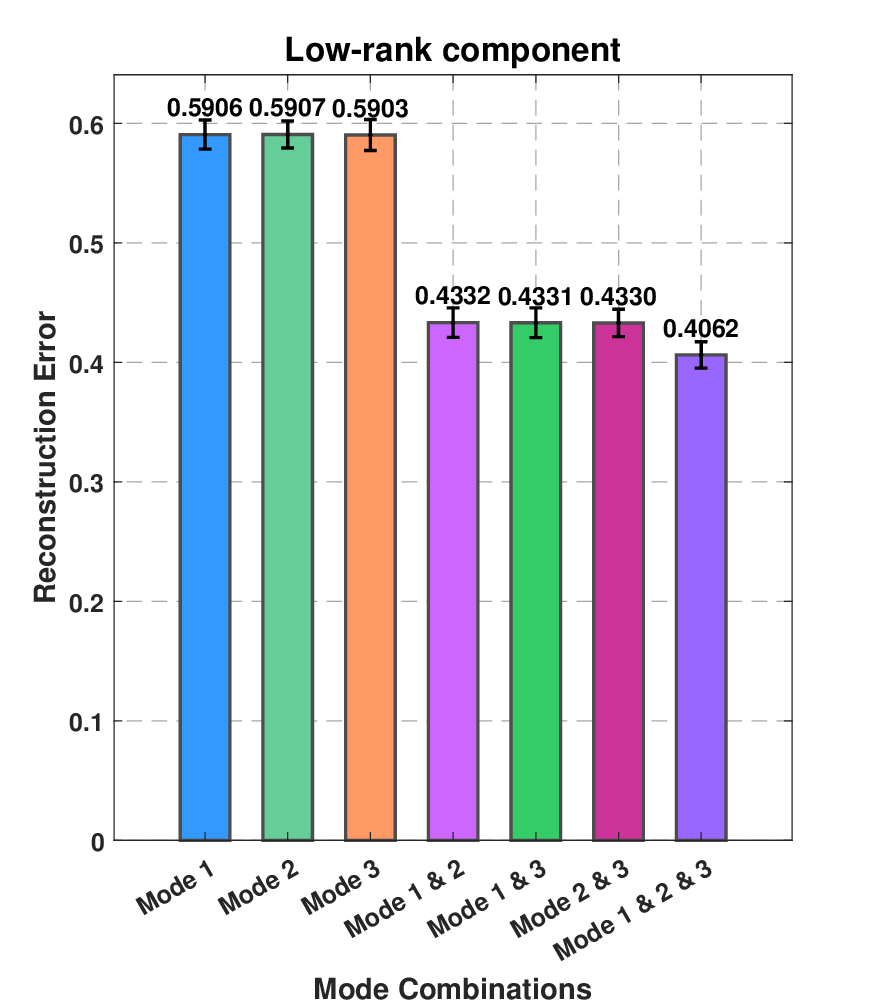}
        \label{fig:image1}
    \end{minipage}\hspace{-0.6em}
    \begin{minipage}[b]{0.24\textwidth} 
        \centering
        \includegraphics[width=\textwidth]{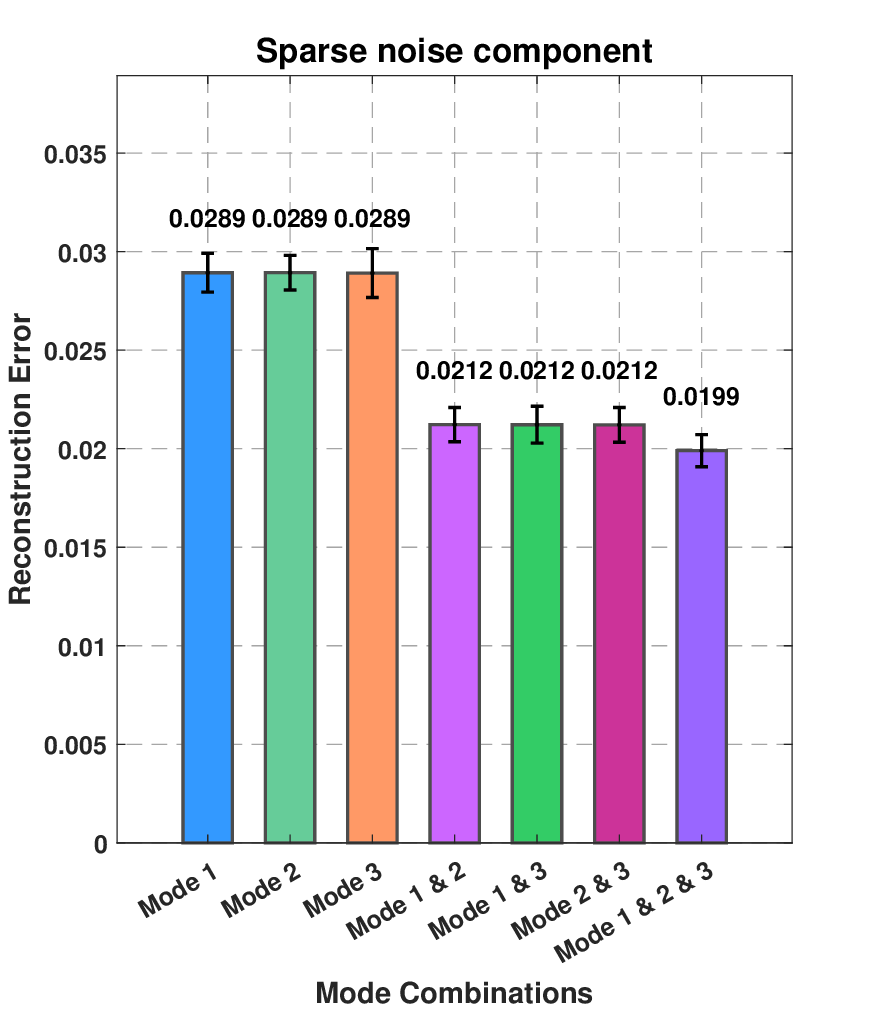}
        \label{fig:image2}
    \end{minipage}

    \caption{Comparison of Reconstruction Errors: Performance of Low-Rank and Sparse Noise Components under Different Mode Combinations.}
    \label{fig2}
\end{figure}

To validate the efficacy of the nuclear norm terms applied to the three mode-permuted tensors in the objective function \eqref{model1}, we design a simple simulation using a three-way tensor dataset with low-rank structure and sparse noise. Specifically, to generate the low-rank tensor $\mathcal{L}\in\mathbb{R}^{n_1\times n_2\times n_3}$, we set the tubal rank of the tensor as $r = 0.1\times n_1$. We create two random low-rank sub-tensors $\mathcal{L}_{11}\in\mathbb{R}^{n_1\times r\times n_3}$ and $\mathcal{L}_{12}\in\mathbb{R}^{r\times n_2\times n_3}$ and construct the first low-rank component $\mathcal{L}_1 = \mathcal{L}_{11}\ast\mathcal{L}_{12}$ using tensor-tensor product (t-product) operations.
Similarly, we generate the second and third low-rank tensors $\mathcal{L}_2 = \mathcal{L}_{21} \ast \mathcal{L}_{22}$ and $\mathcal{L}_3 = \mathcal{L}_{31} \ast \mathcal{L}_{32}$ with $\mathcal{L}_{21}\in\mathbb{R}^{n_1\times r\times n_2}$, $\mathcal{L}_{22}\in\mathbb{R}^{r\times n_3\times n_2}$, $\mathcal{L}_{31}\in\mathbb{R}^{n_3\times r\times n_1}$, $\mathcal{L}_{32}\in\mathbb{R}^{r\times n_2\times n_1}$.
Then, we combine them through permutation operations to form the final low-rank tensor as
\begin{align}
 \mathcal{L} = \mathcal{L}_1 + {\mathcal{L}_{2[2]}} + {\mathcal{L}_{3[3]}}
\end{align}
where ${\mathcal{L}_{2[2]}} = \mathsf{permute}\left(\mathcal{L}_2,[1,3,2]\right)$ and ${\mathcal{L}_{3[3]}} = \mathsf{permute}\left(\mathcal{L}_3,[3,2,1]\right)$.
Additionally, to simulate sparse noise interference, a sparse noise tensor $\mathcal{S} \in\mathbb{R}^{n_1\times n_2\times n_3}$ with a sparsity level of $p=0.05$ is generated and added to the low-rank tensor, resulting in the observation tensor $\mathcal{X} = \mathcal{L} + \mathcal{S}$.

To illustrate the impact of different mode combinations on tensor reconstruction, we conducted experiments using the observation tensor $\mathcal{X}$ under seven settings: (1) Single mode reconstruction (mode 1, mode 2 or mode 3); (2) Pairwise mode combinations (modes 1 \&  2, modes 1 \& 3, or modes 2 \& 3); (3) All mode combinations (modes 1, 2 \&  3). For each setting, we employed the tensor robust principal component analysis (TRPCA) model \cite{8606166} to separate the low-rank and sparse components by minimizing a tailored objective function.

\begin{itemize}
    \item \textit{Single Mode Reconstruction:} When using only one mode, the optimization problem is
    \begin{equation}
        \min_{{\mathcal{L}}, {\mathcal{S}}} \left\|{\mathcal{L}_{[m]}}\right\|_{\circledast} + {\lambda} \left\|{\mathcal{S}}\right\|_1, \quad \text{s.t. } \mathcal{X} = {\mathcal{L}} + {\mathcal{S}},
    \end{equation}
    where $\|\hat{\mathcal{L}}_{[m]}\|_{\circledast}$ computes the tensor nuclear norm along a specific mode, i.e., $\|\hat{\mathcal{L}}\|_{\circledast}$ for mode $m=1$, $\|\hat{\mathcal{L}}_{(1,3,2)}\|_{\circledast}$ for mode $m=2$, or $\|\hat{\mathcal{L}}_{(3,2,1)}\|_{\circledast}$ for mode $m=3$.

    \item \textit{Pairwise Mode Combinations:} When using two modes simultaneously, the optimization problem is
    \begin{equation}
        \min_{{\mathcal{L}}, \mathcal{S}} \left\|{\mathcal{L}}_{[m]}\right\|_{\circledast} + \left\|{\mathcal{L}}_{[j]}\right\|_{\circledast} + {\lambda} \left\|{\mathcal{S}}\right\|_1, \quad \text{s.t. } \mathcal{X} = {\mathcal{L}} + {\mathcal{S}},
    \end{equation}
    where $\|{\mathcal{L}}_{m}\|_{\circledast}$ and $\|{\mathcal{L}}_{j}\|_{\circledast}$ are the nuclear norms along the $m$th and $j$th modes with $(m,j) \in \{(1,2),(1,3),(2,3)\}$.

    \item \textit{All Mode Combinations:} When considering the three modes, the optimization problem is
    \begin{align}
        & \min_{{\mathcal{L}}, {\mathcal{S}}} \left\|{{\mathcal{L}}_{[1]}}\right\|_{\circledast} + \left\|{{\mathcal{L}}_{[2]}}\right\|_{\circledast} + \left\|{{\mathcal{L}}_{[3]}}\right\|_{\circledast} + {\lambda} \left\|{\mathcal{S}}\right\|_1, \nonumber \\
        & \text{s.t. } \mathcal{X} = {{\mathcal{L}}} + {\mathcal{S}}.
    \end{align}
    This formulation jointly enforces low-rankness along the three modes, ensuring comprehensive low-rank representation and expected better reconstruction performance.
\end{itemize}

To quantify the reconstruction performance of each mode combination, the low-rank reconstruction error and sparse component recovery error for each setting are computed as
\begin{align}
 L_{\mbox{er}} \defeq \frac{\left\| \mathcal{L} - \hat{\mathcal{L}} \right\|_\mathsf{F}}{\left\| \mathcal{L} \right\|_\mathsf{F}}, \quad S_{\mbox{er}} \defeq \frac{\left\| \mathcal{S} - \hat{\mathcal{S}} \right\|_\mathsf{F}}{\left\|\mathcal{S}\right\|_\mathsf{F}}
\end{align}
Each experiment is repeated 100 times for each mode setting, and the mean reconstruction errors and standard deviations are computed. Figure \ref{fig2} shows that the reconstruction error for the all mode combination is generally lower than that for any single mode or any pair of combined modes. It indicates that combining low-tubal-rank information from all modes leads to a more comprehensive recovery of the original low-rank structure and better robustness in separating sparse noise components.

\begin{figure}[!t]
    \centering
    \begin{minipage}[b]{0.24\textwidth} 
        \centering
        \includegraphics[width=\textwidth]{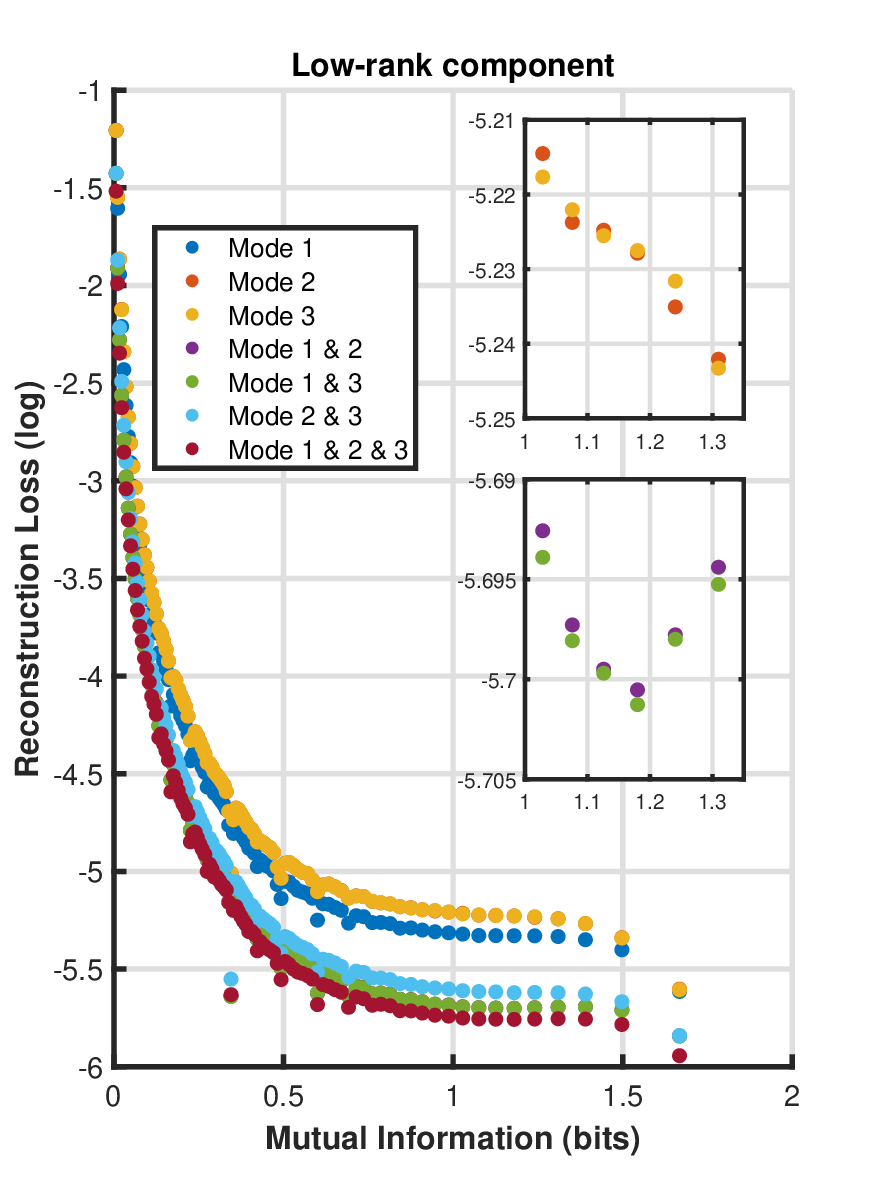}
        \label{fig:image3}
    \end{minipage}\hspace{-0.6em}
    \begin{minipage}[b]{0.24\textwidth} 
        \centering
        \includegraphics[width=\textwidth]{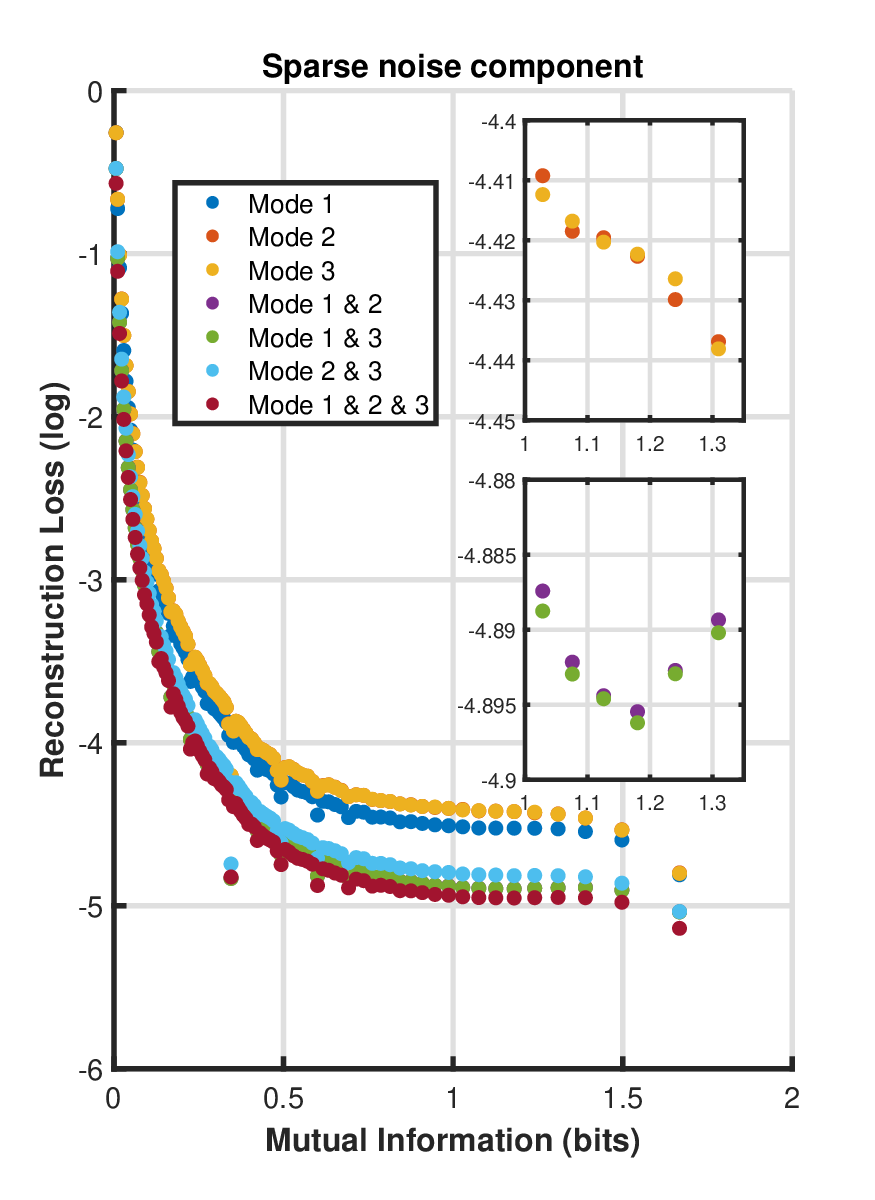}
        \label{fig:image4}
    \end{minipage}

    \caption{Comparison of Reconstruction Errors under Different Mutual Information.}
    \label{fig3}
\end{figure}

To further investigate the impact of inter-view low-rank structures on reconstruction error, we generate a series of simulated tensors with varying inter-view correlations. To be specific, we first generate $100$ views of the form of $100 \times 100$ matrices whose elements are independently drawn from a uniform distribution over (0,1). A set of $99$ tensors with varying levels of inter-view correlation are constructed as follows:
\begin{enumerate}
  \item The first tensor consists of slices that are entirely independent across views by stacking 100 random matrices as different views.
  \item For the second tensor, each slice is constructed as a linear combination of two adjacent matrices from the $100$ generated matrices.
  \item Similarly, for the third tensor, each slice is formed by linearly combining three adjacent matrices, and so on.
  \item In the $99$th tensor, each slice is formed by the linear combination of all the generated matrices except for the one that is farthest in distance.
\end{enumerate}
To quantify the pairwise similarity between the generated random matrices, mutual information (MI) was calculated for all possible matrix pairs:
\begin{align}
I(\mathcal{X})  =  \frac{V(V-1)}{2}  \sum_{v_1 \neq v_2}  p(\mathcal{X}_{::v_1},\mathcal{X}_{::v_2}) \log \frac{p(\mathcal{X}_{::v_1},\mathcal{X}_{::v_2})}{p(\mathcal{X}_{::v_1})p(\mathcal{X}_{::v_2})},
\end{align}
where $p(\mathcal{X}_{::v_1})$ and $p(\mathcal{X}_{::v_2})$ are the marginal distributions of different slice matrices in the tensor $\mathcal{X}$, and $p(\mathcal{X}_{::v_1},\mathcal{X}_{::v_2})$ is the joint distribution of $\mathcal{X}_{::v_1}$ and $\mathcal{X}_{::v_2}$.

\begin{figure*}[htbp]
    \centering

    \begin{tikzpicture}
        \node[anchor=south west, inner sep=0] (title1) at (-6.5, 0) {$p=0.1$};
        \node[anchor=south west, inner sep=0] (title2) at (-3.0, 0) {$p=0.3$};
        \node[anchor=south west, inner sep=0] (title3) at (0.5, 0) {$p=0.5$};
        \node[anchor=south west, inner sep=0] (title4) at (3.75, 0) {$p=0.7$};
        \node[anchor=south west, inner sep=0] (title5) at (7.25, 0) {$p=0.9$};
    \end{tikzpicture}

    \begin{tikzpicture}[baseline]

        \node[inner sep=0] (col1) at (0.0, 0) {
            \begin{minipage}[b]{0.19\textwidth}
                \includegraphics[width=\textwidth]{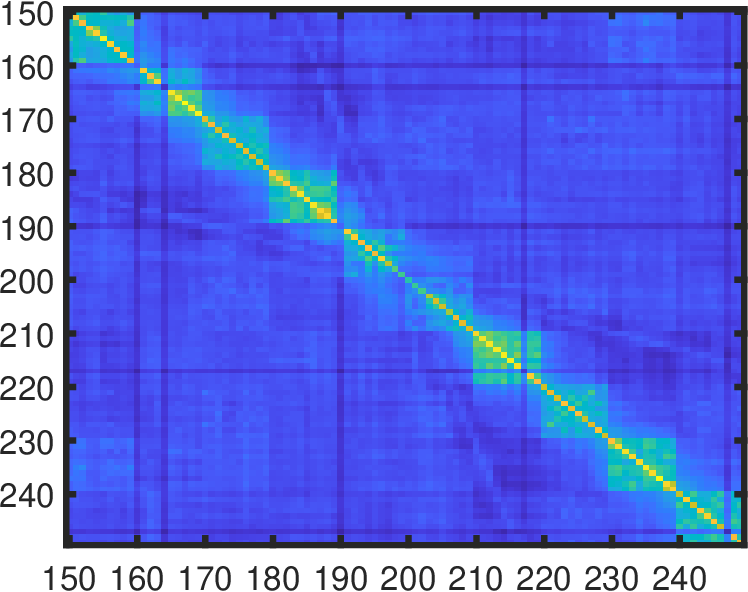} \\[-0.5em]
                \includegraphics[width=\textwidth]{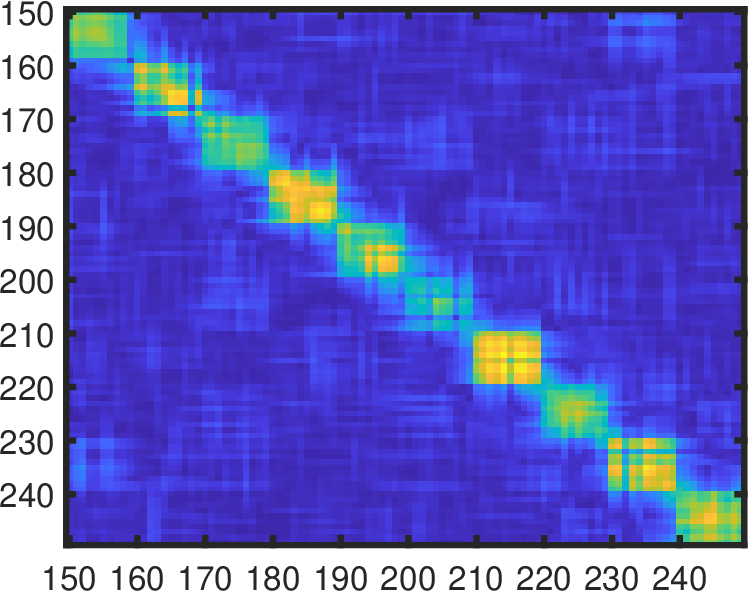} \\[-0.5em]
                \includegraphics[width=\textwidth]{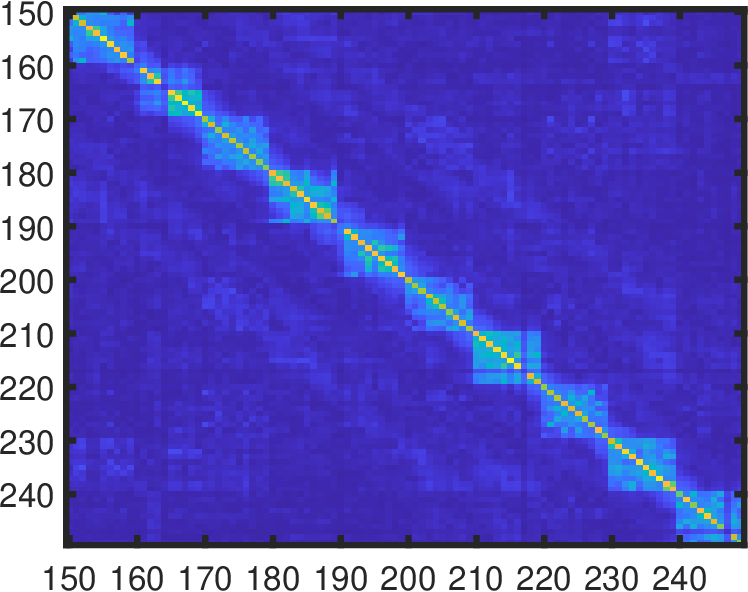}
            \end{minipage}
        };

        \node[inner sep=0] (col2) at (0.19\textwidth, 0) {
            \begin{minipage}[b]{0.19\textwidth}
                \includegraphics[width=\textwidth]{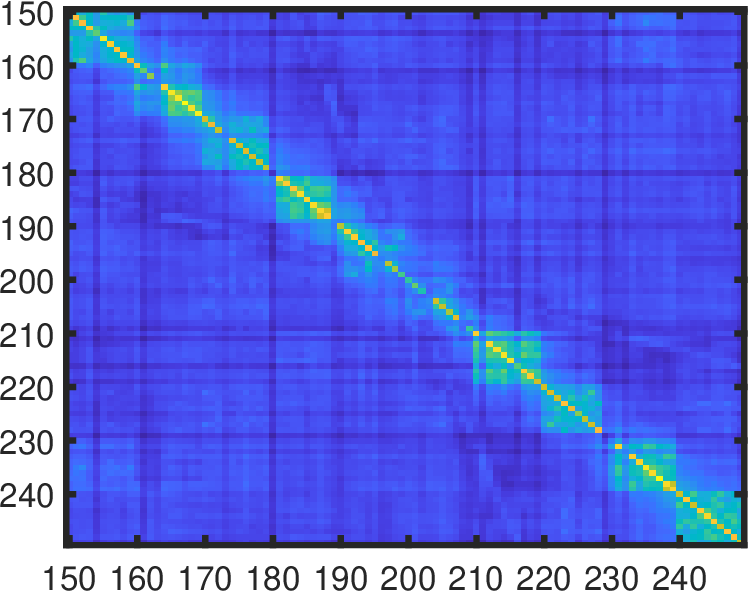} \\[-0.5em]
                \includegraphics[width=\textwidth]{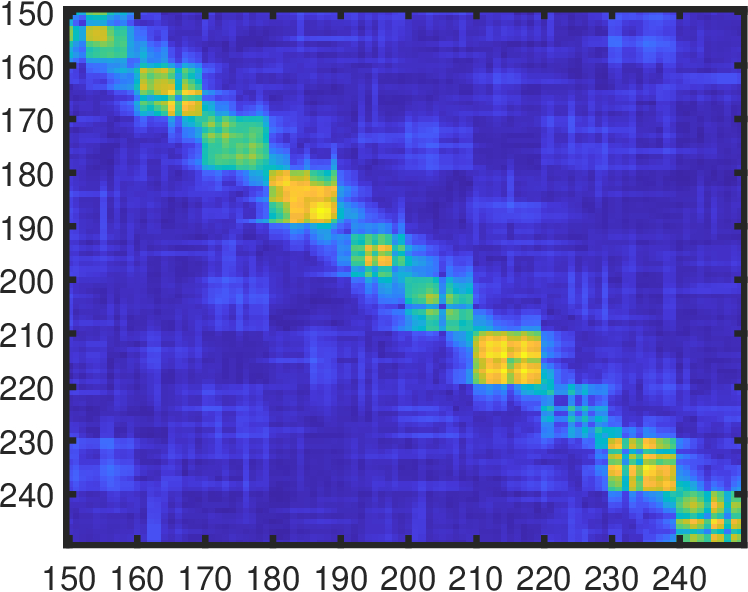} \\[-0.5em]
                \includegraphics[width=\textwidth]{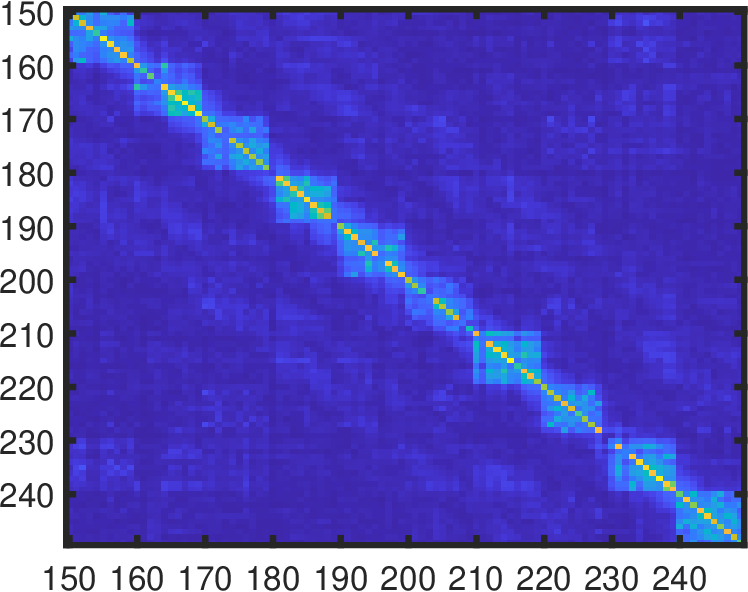}
            \end{minipage}
        };

        \node[inner sep=0] (col3) at (0.38\textwidth, 0) {
            \begin{minipage}[b]{0.19\textwidth}
                \includegraphics[width=\textwidth]{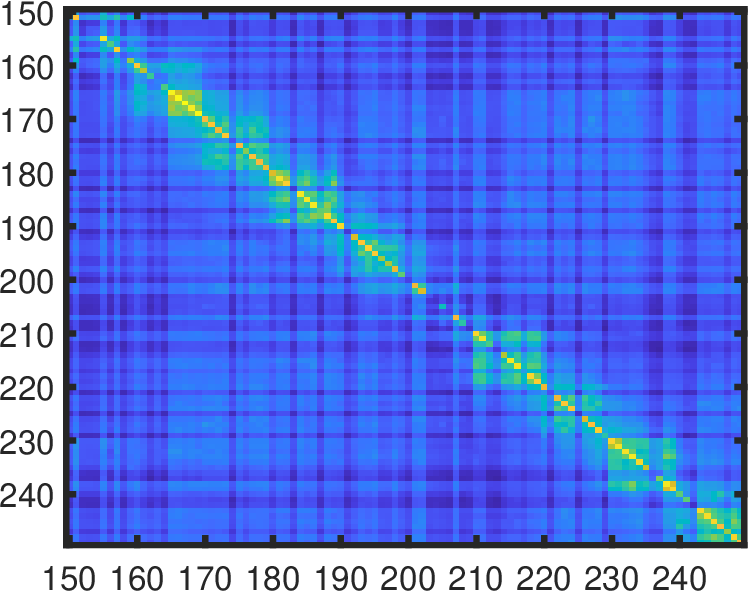} \\[-0.5em]
                \includegraphics[width=\textwidth]{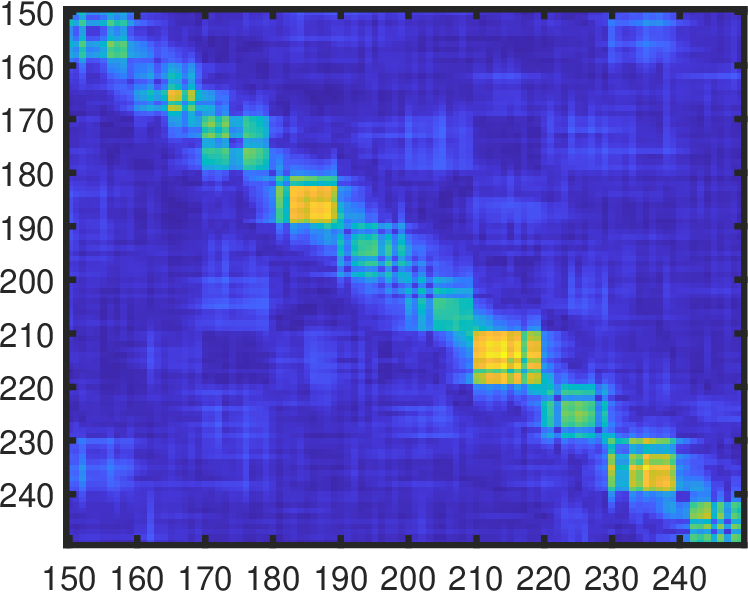} \\[-0.5em]
                \includegraphics[width=\textwidth]{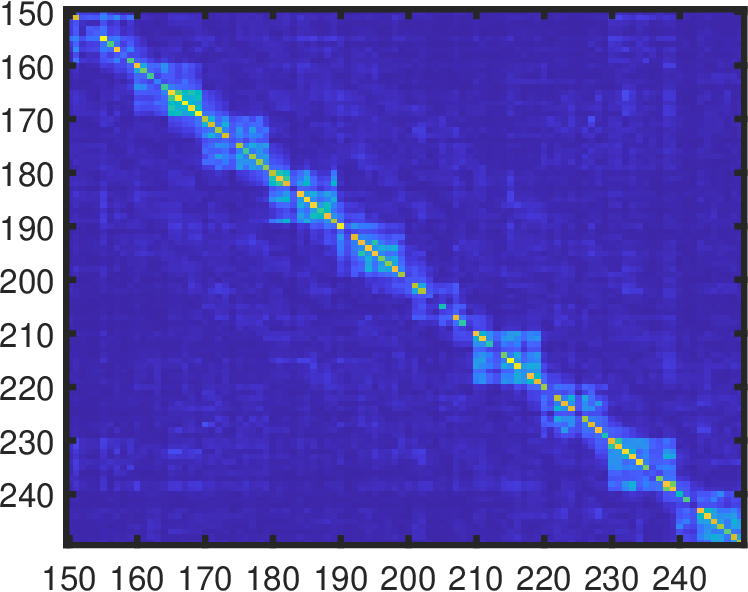}
            \end{minipage}
        };

        \node[inner sep=0] (col4) at (0.57\textwidth, 0) {
            \begin{minipage}[b]{0.19\textwidth}
                \includegraphics[width=\textwidth]{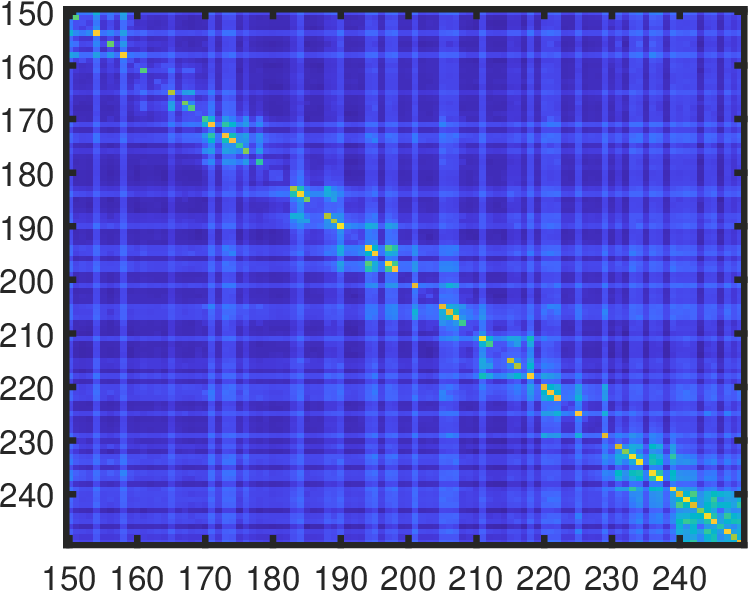} \\[-0.5em]
                \includegraphics[width=\textwidth]{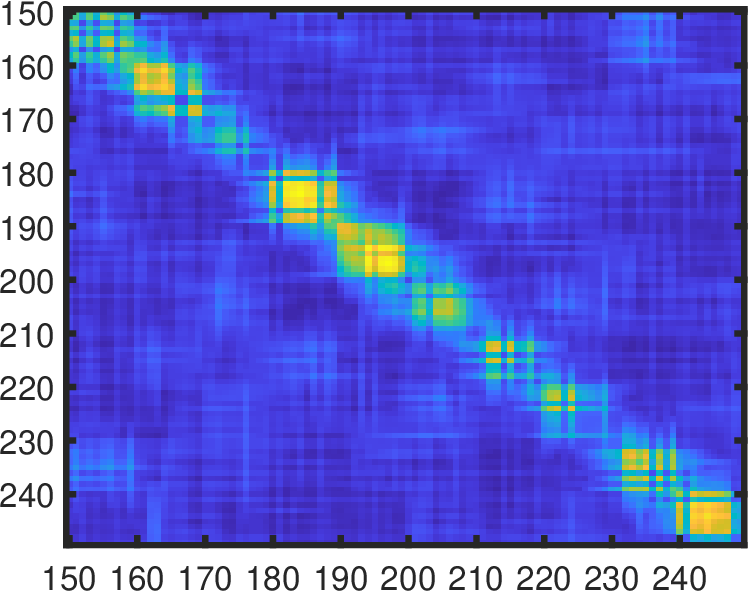} \\[-0.5em]
                \includegraphics[width=\textwidth]{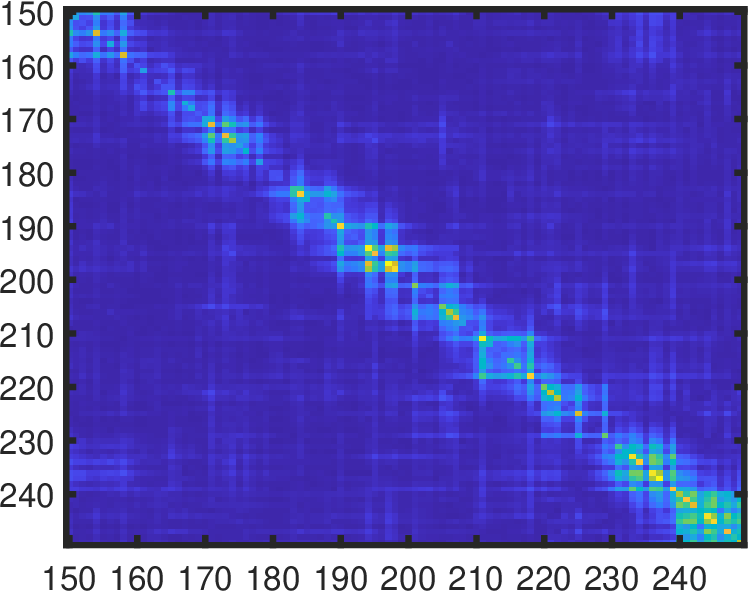}
            \end{minipage}
        };

        \node[inner sep=0] (col5) at (0.76\textwidth, 0) {
            \begin{minipage}[b]{0.19\textwidth}
                \includegraphics[width=\textwidth]{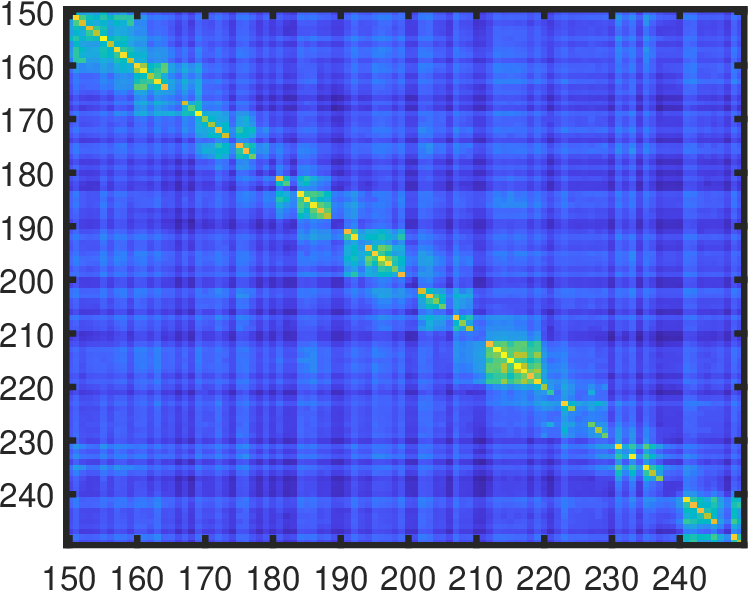} \\[-0.5em]
                \includegraphics[width=\textwidth]{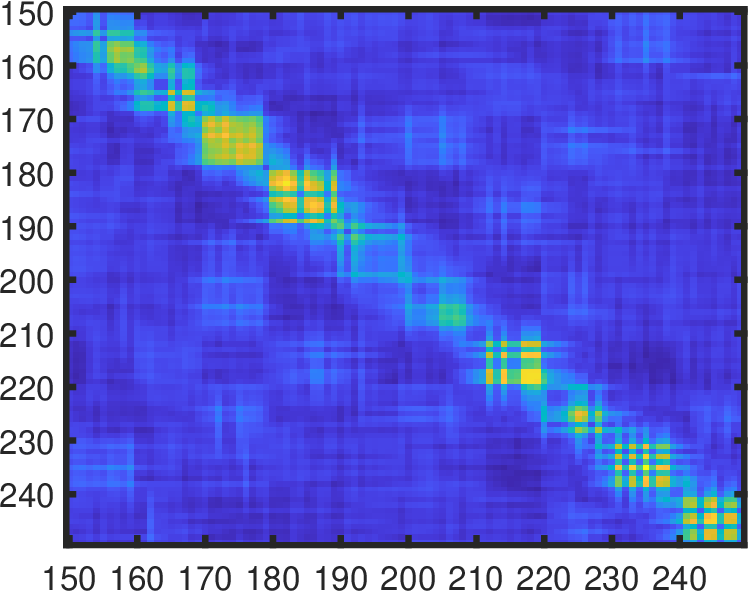} \\[-0.5em]
                \includegraphics[width=\textwidth]{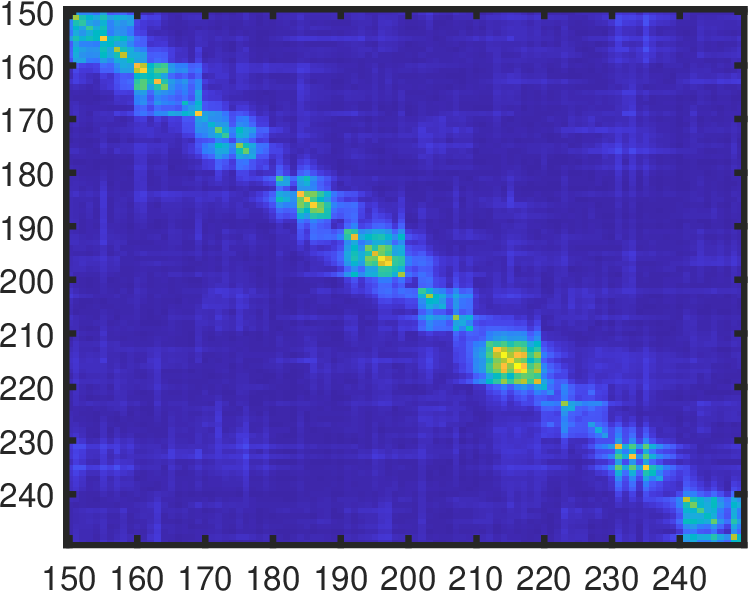}
            \end{minipage}
        };

    \end{tikzpicture}

    \caption{ORL data: visualization of the graphs recovered by ETLSRR (top row), JPLTD (middle row), and the proposed JTIV-LRR method (bottom row) under various missing rates $p$.}
    \label{fig4}
\end{figure*}
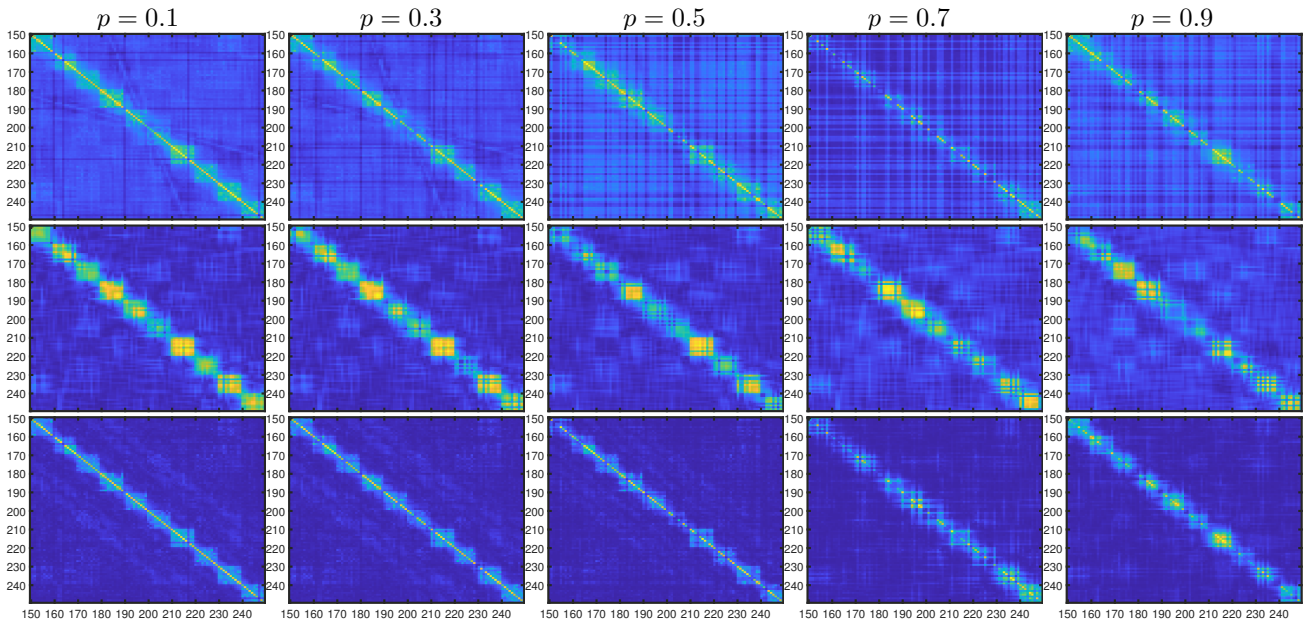

Figure \ref{fig3} demonstrates the relationship between MI and reconstruction loss for both the low-rank component (left panel) and the sparse noise component (right panel). The results highlight the impact of inter-view correlations, as quantified by MI, on the reconstruction performance under various mode combinations of tensor slices.
A clear inverse relationship is observed between MI and reconstruction loss for both components, confirming that the introduced permuted tensor nuclear norm regularizations can effectively capture inter-view correlations, thereby enhancing reconstruction quality, particularly in scenarios with high inter-view correlation. Furthermore, the superior performance of multi-mode combinations compared to single-mode configurations, especially when MI is high, validates the effectiveness of integrating information across multiple views.
These findings not only emphasize the critical role of inter-view correlations in tensor reconstruction but also substantiate the motivation behind this study: leveraging multi-mode information enhances reconstruction accuracy and effectively captures the inherent structure of multiview data.

\subsection{Experimental configuration}

\noindent \textbf{Datasets --} The proposed algorithm JTIV-LRR is then evaluated on seven widely used multi-view datasets, detailed in what follows and summarized in Table \ref{tab:datasets}:
\begin{itemize}
  \item $3$Sources\footnote{\url{http://mlg.ucd.ie/datasets/3sources.html}}: The dataset comprises $984$ news articles drawn from three sources, namely Reuters, BBC, and The Guardian. For the subsequent experiments, $169$ samples are selected as the gallery. These samples are categorized into six distinct topic labels: sports, business, health, technology, entertainment, and politics.
  \item $100$Leaves\footnote{\url{https://archive.ics.uci.edu/dataset/241/one+hundred+plant+species+leaves+data+set}}: The dataset consists of $1600$ samples representing $100$ planetary species. Each sample is characterized using three feature representations, namely shape descriptors, fine-scale margins, and texture histograms.
  \item ORL\footnote{\url{https://cam-orl.co.uk/facedatabase.html}}: The dataset comprises $400$ samples from $40$ distinct individuals. Each face image is represented in three modalities, namely intensity, Local Binary Patterns (LBP), and Gabor features.
  \item ProteinFold\footnote{\url{http://mkl.ucsd.edu/dataset/protein-fold-prediction}} \cite{damoulas2008probabilistic}: The dataset comprises $694$ samples for protein fold prediction, represented through multi-kernel learning feature embeddings. Each sample encompasses $12$ distinct feature views.
  \item Scene\footnote{\url{http://www-cvr.ai.uiuc.edu/ponce_grp/data/}} \cite{fei2005bayesian}: The dataset comprises $4485$ images spanning $15$ distinct categories of indoor and outdoor scenes. Following the methodology outlined in \cite{dai2013ensemble}, we extract $20$-dimensional GIST features and $50$-dimensional PHOG features for each image.
  \item Mfeat\footnote{\url{https://archive.ics.uci.edu/dataset/72/multiple+features}}: The Multiple Features (Mfeat) dataset consists of handwritten digit samples (0–9), each characterized by six complementary feature representations: Fourier descriptors of character contours, profile correlation metrics, Karhunen-Loève coefficients, pixel intensity averages computed over $2\times 3$ sliding windows, Zernike moment invariants, and a set of six morphological attributes.
  \item Caltech101-7\footnote{\url{https://data.caltech.edu/records/mzrjq-6wc02}} \cite{1384978}: This dataset, derived from a subset of Caltech101, contains $1474$ samples spanning seven object categories. Each sample is characterized by six distinct feature representations: Gabor filter responses, wavelet moment descriptors, CENTRIST features, Histogram of Oriented Gradients (HOG), GIST embeddings, and Local Binary Patterns (LBP).
\end{itemize}
It is worth  noting that these multi-view datasets are originally complete across all views. To simulate an incomplete scenario, we randomly select $n_s$ samples and remove their corresponding data in one or multiple views, while keeping data from at least one view remains available. The missing ratios ($p = \frac{n_s}{n}$) are set to $0.1$, $0.3$, $0.5$, $0.7$, and $0.9$, respectively.\\

\begin{figure*}[htbp]
    \centering

    \begin{tikzpicture}
        \node[anchor=south west, inner sep=0] (title1) at (-6.5, 0) {$p=0.1$};
        \node[anchor=south west, inner sep=0] (title2) at (-3.0, 0) {$p=0.3$};
        \node[anchor=south west, inner sep=0] (title3) at (0.5, 0) {$p=0.5$};
        \node[anchor=south west, inner sep=0] (title4) at (3.75, 0) {$p=0.7$};
        \node[anchor=south west, inner sep=0] (title5) at (7.25, 0) {$p=0.9$};
    \end{tikzpicture}

    \begin{tikzpicture}[baseline]

        \node[inner sep=0] (col1) at (0.0, 0) {
            \begin{minipage}[b]{0.19\textwidth}
                \includegraphics[width=\textwidth]{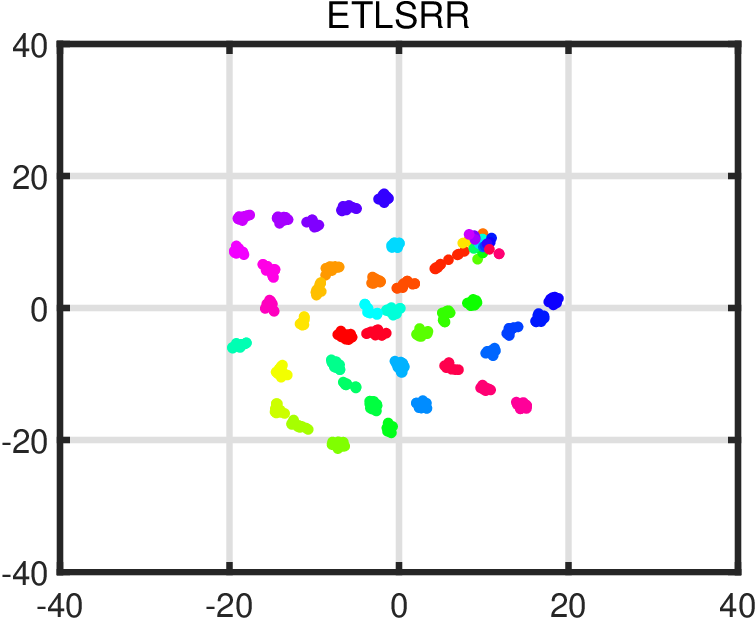} \\[0.65em]
                \includegraphics[width=\textwidth]{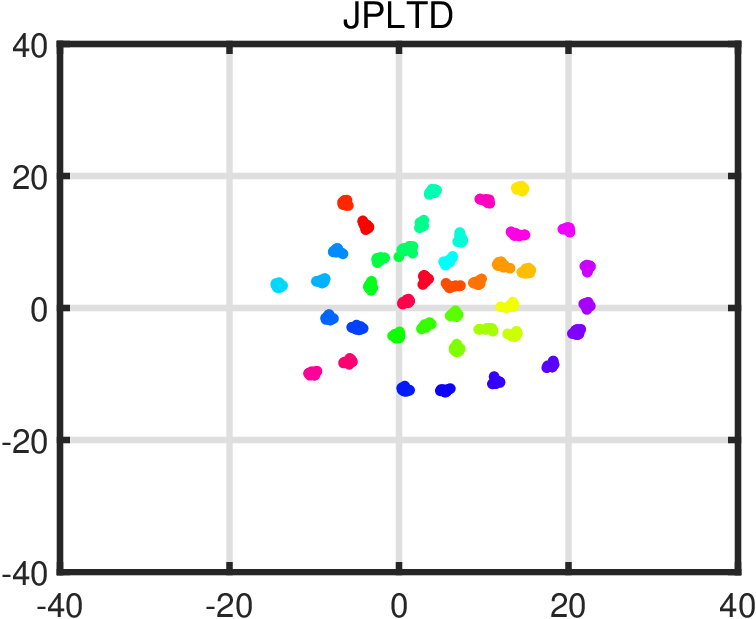} \\[0.65em]
                \includegraphics[width=\textwidth]{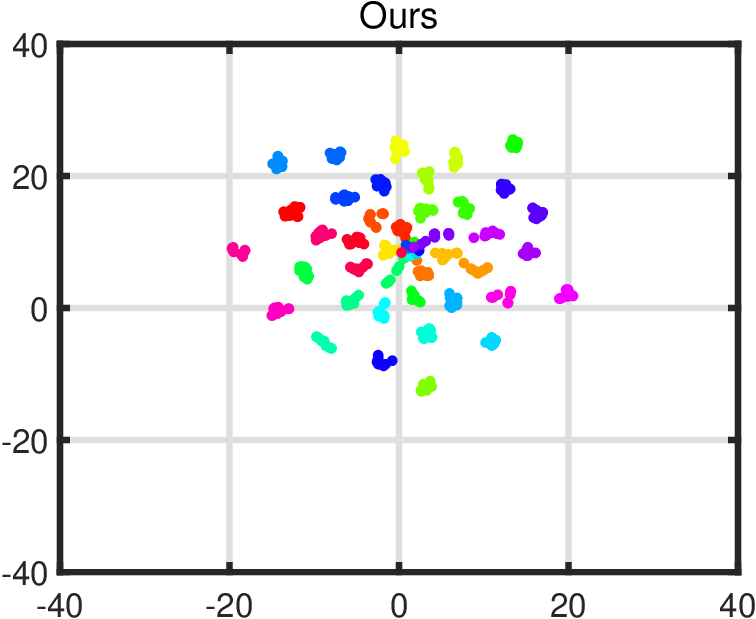}
            \end{minipage}
        };

        \node[inner sep=0] (col2) at (0.19\textwidth, 0) {
            \begin{minipage}[b]{0.19\textwidth}
                \includegraphics[width=\textwidth]{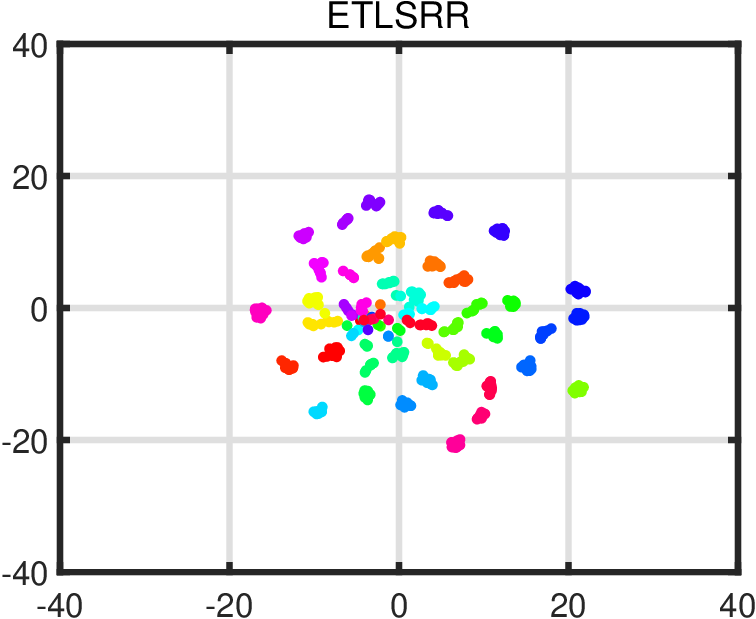} \\[0.65em]
                \includegraphics[width=\textwidth]{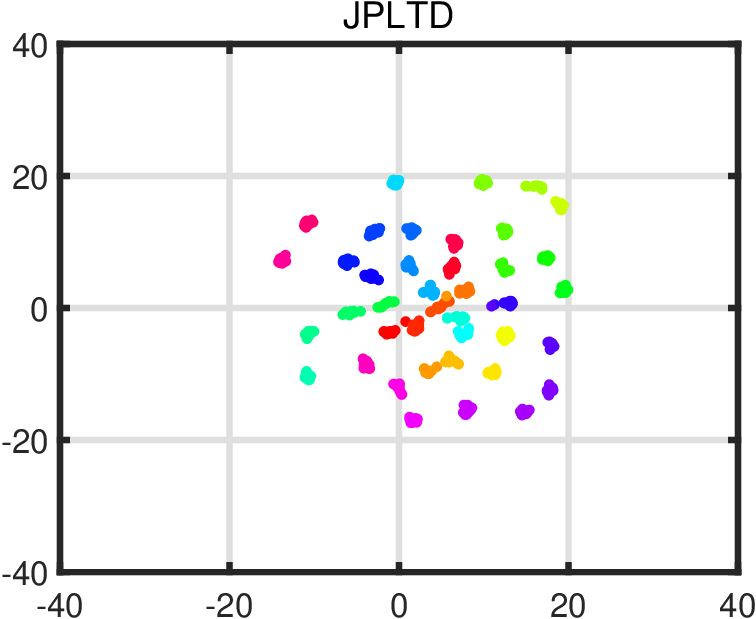} \\[0.65em]
                \includegraphics[width=\textwidth]{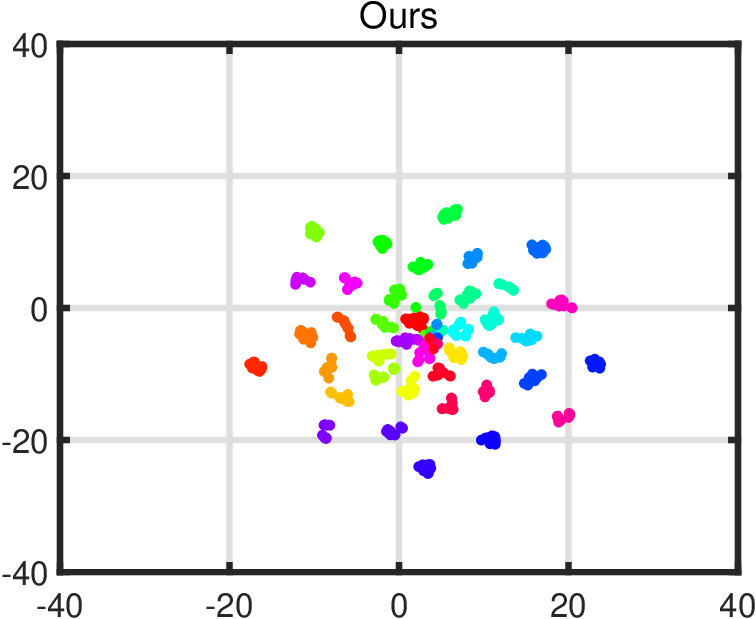}
            \end{minipage}
        };

        \node[inner sep=0] (col3) at (0.38\textwidth, 0) {
            \begin{minipage}[b]{0.19\textwidth}
                \includegraphics[width=\textwidth]{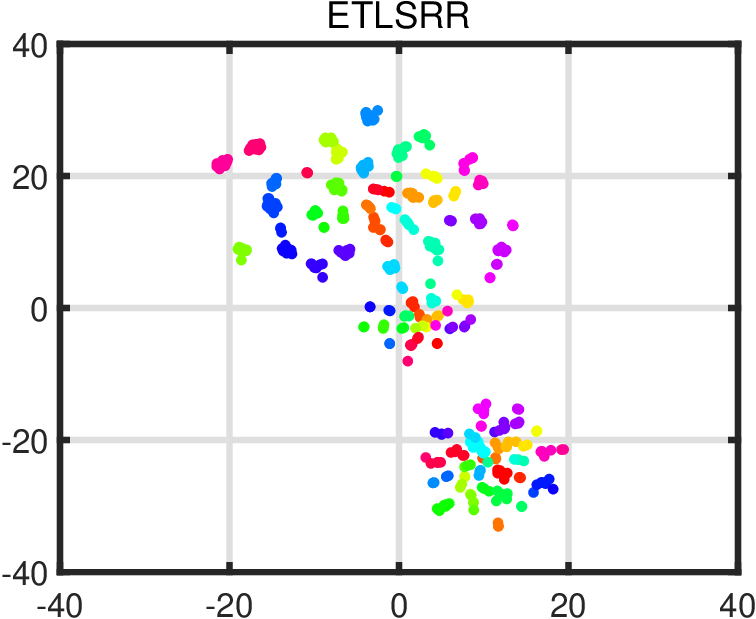} \\[0.65em]
                \includegraphics[width=\textwidth]{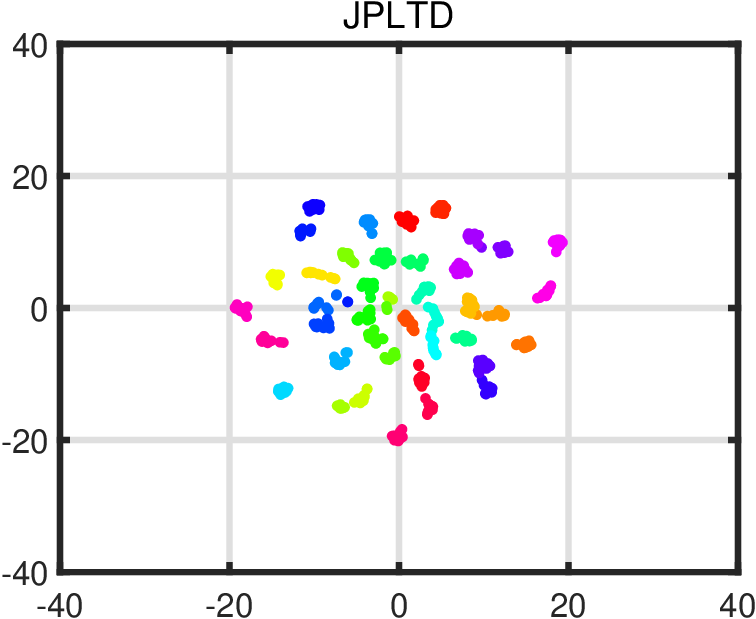} \\[0.65em]
                \includegraphics[width=\textwidth]{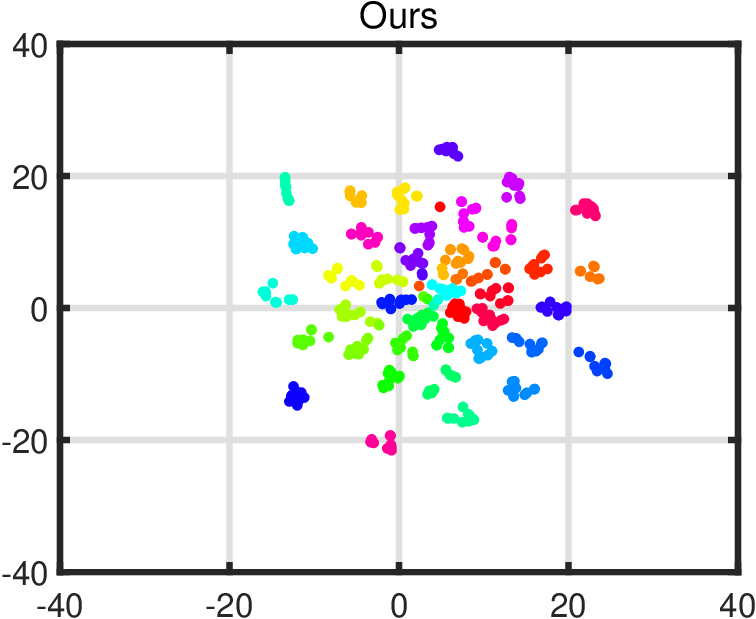}
            \end{minipage}
        };

        \node[inner sep=0] (col4) at (0.57\textwidth, 0) {
            \begin{minipage}[b]{0.19\textwidth}
                \includegraphics[width=\textwidth]{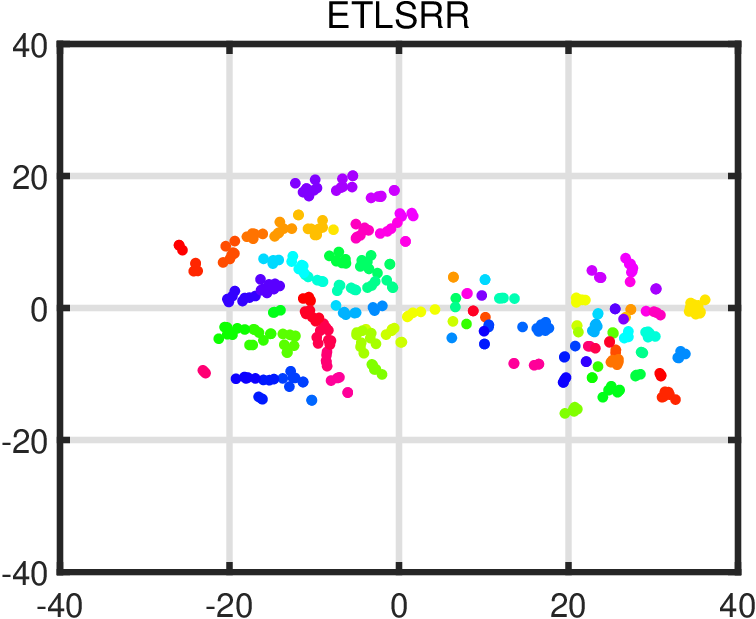} \\[0.65em]
                \includegraphics[width=\textwidth]{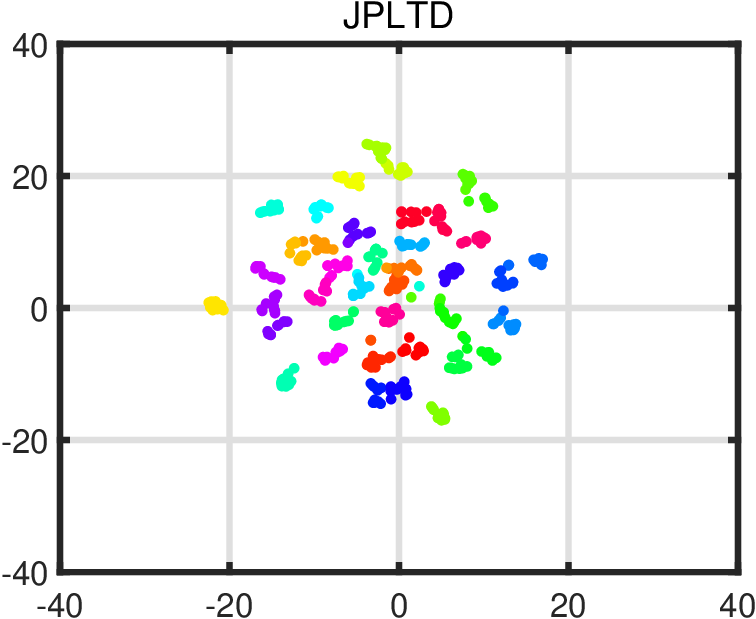} \\[0.65em]
                \includegraphics[width=\textwidth]{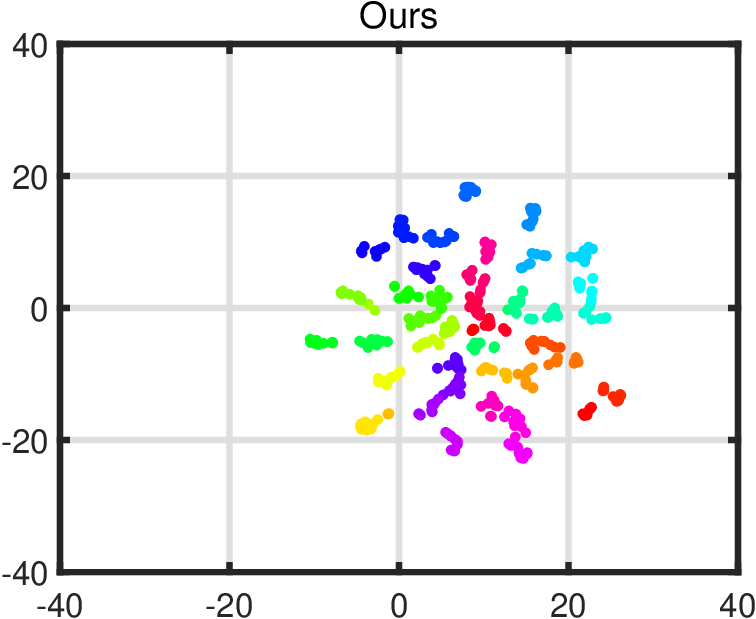}
            \end{minipage}
        };

        \node[inner sep=0] (col5) at (0.76\textwidth, 0) {
            \begin{minipage}[b]{0.19\textwidth}
                \includegraphics[width=\textwidth]{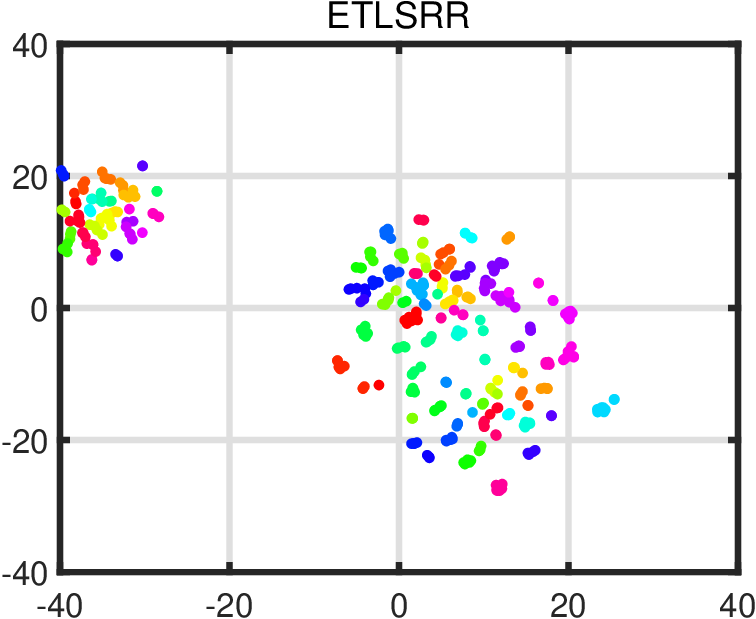} \\[0.65em]
                \includegraphics[width=\textwidth]{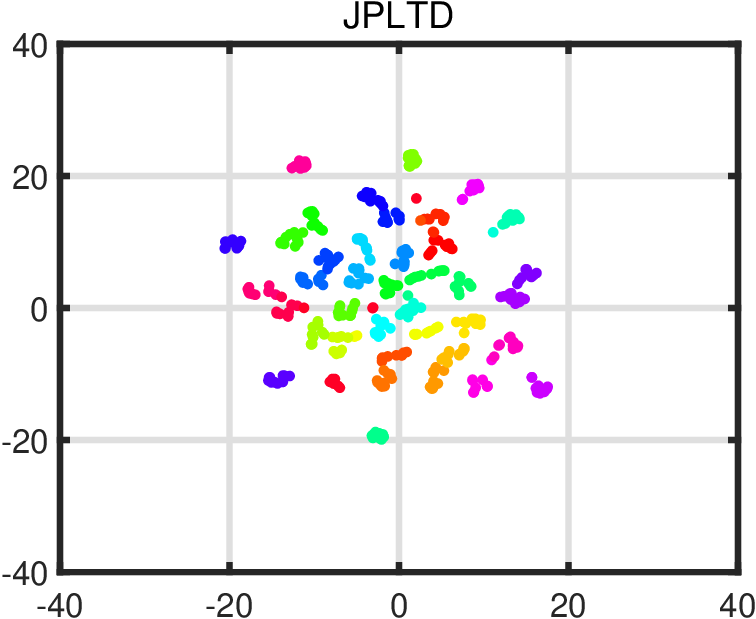} \\[0.65em]
                \includegraphics[width=\textwidth]{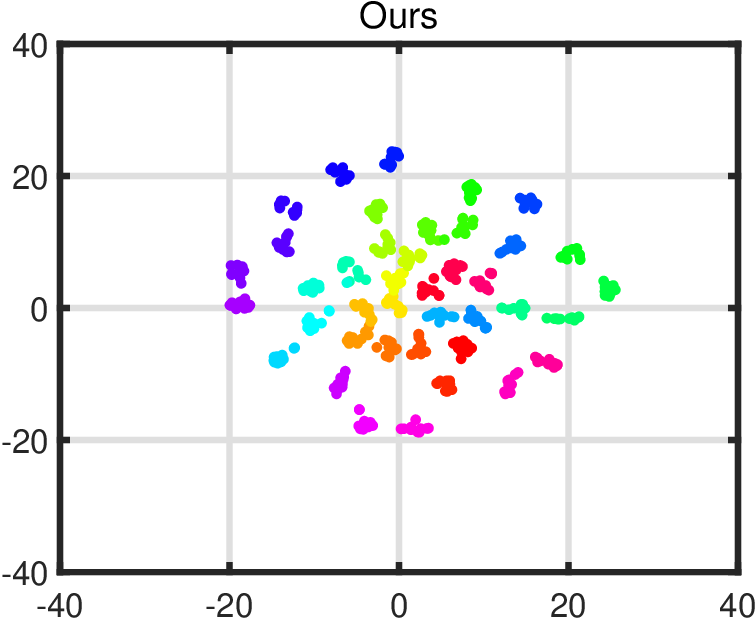}
            \end{minipage}
        };

    \end{tikzpicture}

    \caption{Enhanced t-SNE visualization of recovered graphs for ETLSRR, JPLTD, and our method on ORL data across different missing rate $p$.}
    \label{fig5}
\end{figure*}
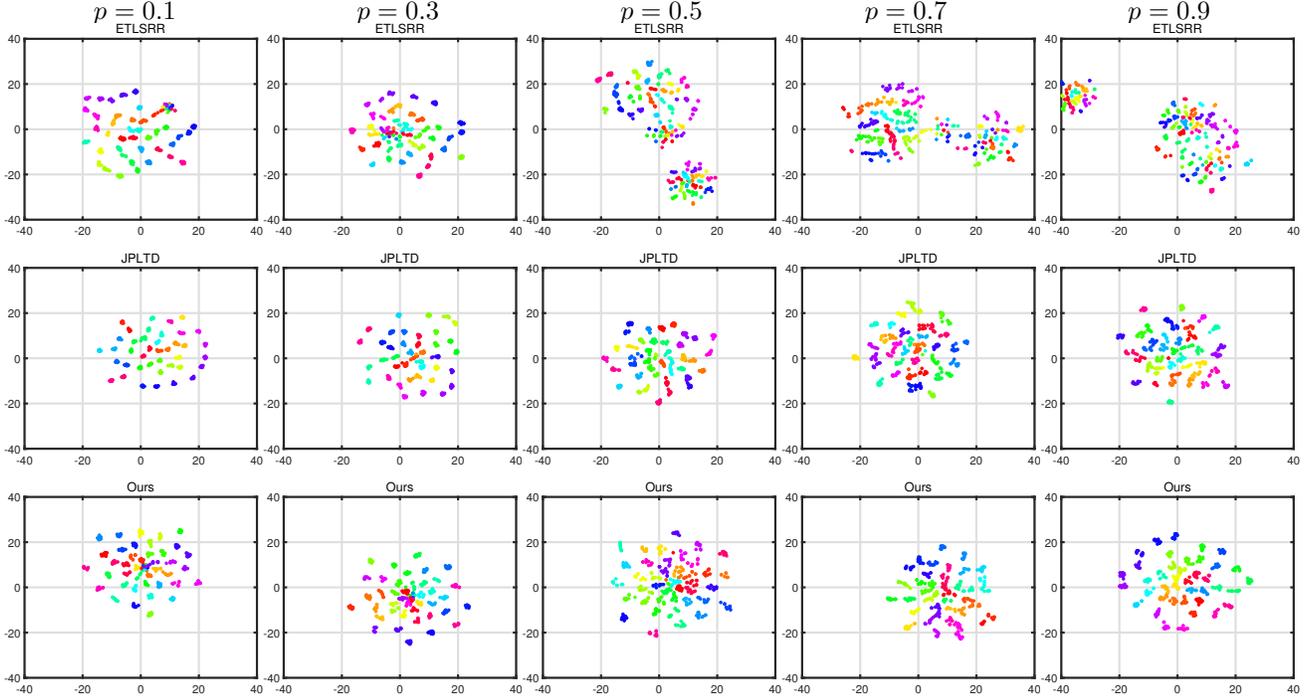

\noindent \textbf{Compared methods -- } To emphasize the advantages of the proposed JTIV-LRR method, seven representative incomplete multi-view clustering approaches are selected as benchmark competitors.
In our experiments, the hyperparameters are set to ${\lambda_1} = 10,\ {\lambda_2} = 10,\ \textrm{and},\ {\lambda_3} = 10$. For the parameter sensitivity analysis, we evaluate these parameters over a range of values: $\{ 1, 2, 3, 4, 5, 6, 7, 8, 9, 10, 20, 30, 40, 50, 100 \}$.
\begin{itemize}
  \item TMBSD \cite{9428106} investigates the block-diagonal structures of incomplete multiview data under tensor low-rank constraints. In our experiments, the default values of the two regularization parameters $\lambda_1 = 10, \lambda_2 = 10^{-3}$ are adopted.
  \item TIMC \cite{wen2021unified} recovers missing views and leverages complete and intra-view information to improve clustering performance. The default parameters $\lambda_1 = 10^{-5}, \lambda_2 = 5*10^{-3}, \lambda_3 = 5*10^{-3}$ are adopted.
  \item TCIMC \cite{xia2022tensor} uses tensor Schatten $p$-norm-based completion to integrate inter-view and intra-view similarities, ensuring low-rank structure and connectivity for effective clustering. The default parameters, $p = 0.6\ \textrm{and}\ \beta = 0.3$, are adopted.
  \item LATER \cite{10004588} recovers missing views from a latent representation, learns multilevel graphs, and employs tensor nuclear norm regularization for enhanced clustering. The default parameters $\lambda_1 = 1, \lambda_2 = 0.01, \lambda_3 = 0.1$ are adopted.
  \item JPLTD \cite{lv2023joint} reduces feature redundancy and noise via orthogonal projection, learns low-dimensional similarity graphs, and uses tensor decomposition for robust clustering. The default parameters $\lambda=5, \theta = 0.1$ are adopted.
  \item IMVCCBG \cite{wang2022highly} applies a bipartite graph framework with multi-view anchor learning, achieving linear complexity for large-scale clustering tasks.
  \item ETLSRR \cite{zhang2023enhanced}: Combines incomplete graph learning with tensor decomposition, using nonconvex low-rank regularization to reduce noise and enhance clustering. The default parameters $\mu = 1, \lambda = 10, \theta = 1, \gamma = 20$ are adopted.
\end{itemize}

\noindent \textbf{Model evaluation metrics --}
Clustering accuracy (ACC), normalized mutual information (NMI), and adjusted rand index (ARI) are employed as evaluation metrics. For all metrics, higher values indicate better performance. All experiments are repeated for 10 times, and the mean values of these metrics are reported.

\begin{table*}[ht]
\centering
\caption{Comparison of different methods across various datasets with different missing rate $p$. Red bold font indicates the best performance under the current condition, while blue bold font represents the second-best performance.}
\begin{adjustbox}{max width=\textwidth}
\begin{tabular}{|c|c|ccc|ccc|ccc|ccc|ccc|ccc|}
\hline
\multirow{2}{*}{$P$} & \multirow{2}{*}{Metric} & \multicolumn{3}{c|}{0.1} & \multicolumn{3}{c|}{0.3} & \multicolumn{3}{c|}{0.5} & \multicolumn{3}{c|}{0.7} & \multicolumn{3}{c|}{0.9}\\
\cline{3-17}
 & & ACC & NMI & ARI & ACC & NMI & ARI & ACC & NMI & ARI & ACC & NMI & ARI & ACC & NMI & ARI \\
\hline
\multirow{9}{*}{3 sources}
& \cellcolor{lightpurple}TMBSD   & 49.93 & 37.95 & 28.17 & 49.15 & 34.64 & 25.65 & 46.12 & 30.81 & 22.25 & 43.18 & 24.98 & 17.63 & 42.66 & 25.46 & 17.24 \\
& \cellcolor{lightpurple}TIMC    & 55.11 & 57.02 & 44.59 & 51.30 & 48.72 & 36.80 & 48.49 & 37.69 & 28.28 & 38.31 & 22.06 & 14.50 & 32.76 & 15.50 & 08.95 \\
& \cellcolor{lightpurple}TCIMC   & 43.31 & 31.25 & 17.65 & 44.61 & 34.54 & 19.21 & 45.75 & 36.99 & 20.34 & 47.46 & 38.18 & 20.94 & 45.27 & 38.21 & 19.97 \\
& \cellcolor{lightpurple}LATER   & 67.18 & 61.28 & 51.80 & \bluebold{67.38} & 60.19 & 50.93 & 65.25 & 59.37 & 49.60 & 62.56 & 55.11 & 45.42 & 58.89 & 47.66 & 41.24 \\
& \cellcolor{lightpurple}JPLTD   & 66.00 & 58.91 & 51.02 & 66.86 & 59.72 & 52.01 & 60.69 & 57.40 & 46.57 & 60.48 & 55.45 & 46.36 & 58.82 & 49.88 & 42.17 \\
& \cellcolor{lightpurple}IMVCCBG & 65.22 & \redbold{82.37} & \bluebold{54.19} & 66.17 & \redbold{82.90} & \bluebold{55.35} & \bluebold{66.61} & \redbold{83.24} & \redbold{55.98} & \bluebold{67.14} & \redbold{83.51} & \redbold{55.65} & \redbold{67.24} & \redbold{83.61}\redbold & \redbold{56.83} \\
& \cellcolor{lightpurple}ETLSRR  & \bluebold{67.83} & 61.98 & 52.69 & 60.48 & 57.53 & 45.84 & 57.57 & 55.77 & 42.84 & 60.11 & 55.76 & 45.95 & {62.73} & \redbold{53.29} & {44.15} \\
& \cellcolor{lightpurple}Proposed    & \redbold{79.38} & \bluebold{72.70} & \redbold{65.88} & \redbold{72.24} & \bluebold{60.57} & \redbold{57.03} & \redbold{69.05} & \bluebold{60.03} & \bluebold{53.11} & \redbold{68.75} & \bluebold{56.92} & \bluebold{52.84} & \bluebold{63.72} & {51.54} & \bluebold{45.24} \\
\hline
\multirow{9}{*}{100 leaves}
& \cellcolor{lightpurple}TMBSD   & 86.98 & 94.07 & 82.80 & 86.79 & 94.05 & 82.73 & 86.64 & 94.22 & 82.92 & \bluebold{87.38} & \bluebold{94.64} & \bluebold{84.04} & \bluebold{87.21} & \redbold{94.46} & \bluebold{83.62} \\
& \cellcolor{lightpurple}TIMC    & 83.90 & 93.31 & 80.02 & 73.96 & 86.37 & 63.89 & 62.13 & 79.01 & 48.04 & 48.36 & 71.14 & 32.02 & 41.13 & 67.25 & 24.98 \\
& \cellcolor{lightpurple}TCIMC   & 77.89 & 89.05 & 70.36 & 74.81 & 86.42 & 64.81 & 70.71 & 83.40 & 58.79 & 64.34 & 79.02 & 50.02 & 58.21 & 75.35 & 42.89 \\
& \cellcolor{lightpurple}LATER   & 89.82 & 96.45 & 88.26 & 86.77 & 94.28 & 83.17 & 79.23 & 89.13 & 70.76 & 64.70 & 80.39 & 50.97 & 53.74 & 74.37 & 38.45 \\
& \cellcolor{lightpurple}JPLTD   & \redbold{91.40} & \redbold{97.17} & \redbold{91.76} & \bluebold{89.40} & \bluebold{95.55} & \redbold{89.00} & \bluebold{88.71} & \bluebold{95.17} & \redbold{87.98} & \redbold{88.44} & \redbold{94.86} & \redbold{87.25} & \redbold{87.59} & \bluebold{94.39} & \redbold{85.95} \\
& \cellcolor{lightpurple}IMVCCBG & 49.99 & 43.84 & 30.75 & 46.50 & 41.57 & 28.20 & 46.34 & 39.65 & 26.13 & 45.99 & 38.47 & 24.85 & 45.54 & 38.28 & 24.54 \\
& \cellcolor{lightpurple}ETLSRR  & 88.11 & 95.86 & 86.36 & 88.06 & 95.53 & 85.77 & 86.82 & 94.50 & 83.46 & 81.86 & 91.12 & 75.45 & 71.32 & 84.82 & 60.91\\
& \cellcolor{lightpurple}Proposed    & \bluebold{90.16} & \bluebold{96.78} & \bluebold{88.93} & \redbold{90.03} & \redbold{96.49} & \bluebold{88.37} & \redbold{89.37} & \redbold{95.75} & \bluebold{86.80} & {85.36} & {93.08} & {80.29} & {75.33} & {87.19} & {66.17}\\
\hline
\multirow{9}{*}{ORL}
& \cellcolor{lightpurple}TMBSD & 95.22 & 98.15 & 94.42 & 91.90 & 95.94 & 88.98 & 90.29 & 95.44 & 87.38 & 90.01 & 95.25 & 86.84 & 89.37 & 95.25 & 86.03\\
& \cellcolor{lightpurple}TIMC & 77.84 & 87.44 & 67.60 & 31.13 & 46.32 & 10.10 & 15.97 & 36.74 & 00.13 & 17.46 & 41.11 & 00.96 & 16.67 & 40.50 & 00.31 \\
& \cellcolor{lightpurple}TCIMC & 74.98 & 86.90 & 64.58 & 75.53 & 87.01 & 65.38 & 74.50 & 86.32 & 63.84 & 72.50 & 84.56 & 60.25 & 71.19 & 83.36 & 58.01 \\
& \cellcolor{lightpurple}LATER & 95.72 & 98.01 & 95.60 & 95.16 & \bluebold{98.79} & 94.92 & 94.87 & 97.60 & 94.44 & 94.11 & 97.25 & \bluebold{93.46} & 92.75 & 96.57 & 91.61 \\
& \cellcolor{lightpurple}JPLTD & \redbold{97.39} & \redbold{99.20} & \redbold{97.25} & \bluebold{96.33} & {98.73} & \bluebold{95.90} & \bluebold{95.80} & \bluebold{98.43} & \bluebold{95.15} & \bluebold{94.18} & \bluebold{97.50} & 92.72 & 92.40 & 96.53 & 90.14 \\
& \cellcolor{lightpurple}IMVCCBG & 79.15 & 89.89 & 72.83 & 79.38 & 89.88 & 72.75 & 79.33 & 89.97 & 72.89 & 78.17 & 89.41 & 71.49 & 77.51 & 88.96 & 70.51 \\
& \cellcolor{lightpurple}ETLSRR & 95.86 & 98.76 & 95.77 & 95.29 & 98.53 & 95.01 & 94.89 & 98.23 & 94.33 & 93.63 & 97.60 & 92.64 & \bluebold{93.00} & \bluebold{97.28} & \bluebold{91.75} \\
& \cellcolor{lightpurple}Proposed & \bluebold{96.28} & \bluebold{98.93} & \bluebold{96.28} & \redbold{97.36} & \redbold{99.07} & \redbold{97.03} & \redbold{96.71} & \redbold{98.80} & \redbold{96.20} & \redbold{95.28} & \redbold{98.14} & \redbold{94.39} & \redbold{94.33} & \redbold{97.70} & \redbold{93.15} \\
\hline
\multirow{9}{*}{ProteinFold}
& \cellcolor{lightpurple}TMBSD & 45.13 & 56.76 & 30.23 & 43.43 & 53.84 & 28.00 & 41.52 & 50.37 & 25.69 & 40.72 & 49.52 & 25.42 & 40.37 & 49.05 & 24.61 \\
& \cellcolor{lightpurple}TIMC & 37.37 & 45.65 & 20.56 & 36.78 & 44.86 & 20.18 & 34.98 & 42.59 & 17.90 & 34.19 & 42.31 & 18.25 & 34.87 & 42.99 & 18.76 \\
& \cellcolor{lightpurple}TCIMC & 27.01 & 36.99 & 11.57 & 26.82 & 37.05 & 11.39 & 29.81 & 39.16 & 13.81 & 28.92 & 37.96 & 12.90 & 27.30 & 36.29 & 11.34 \\
& \cellcolor{lightpurple}LATER & 41.57 & 49.69 & 23.91 & 40.47 & 48.56 & 22.93 & 39.09 & 46.77 & 21.56 & 37.53 & 45.32 & 20.27 & 36.74 & 44.49 & 19.56 \\
& \cellcolor{lightpurple}JPLTD & \redbold{53.80} & \redbold{63.90} & \redbold{38.19} & \redbold{52.12} & \bluebold{61.77} & \redbold{36.39} & \bluebold{50.19} & \bluebold{60.30} & \bluebold{34.39} & \bluebold{48.96} & \bluebold{58.60} & \bluebold{33.39} & \bluebold{46.73} & \bluebold{55.86} & \bluebold{30.46} \\
& \cellcolor{lightpurple}IMVCCBG & 29.75 & 38.80 & 13.66 & 29.72 & 38.79 & 13.56 & 29.42 & 38.71 & 13.50 & 29.78 & 38.89 & 13.65 & 29.71 & 38.89 & 13.57 \\
&  \cellcolor{lightpurple}ETLSRR & 48.10 & 58.32 & 32.38 & 48.25 & 58.66 & 32.95 & 49.08 & 59.09 & 33.54 & 46.87 & 57.10 & 31.41 & 45.19 & 54.23 & 29.06 \\
& \cellcolor{lightpurple}Proposed & \bluebold{50.25} & \bluebold{62.40} & \bluebold{34.90} & \bluebold{50.84} & \redbold{62.84} & \bluebold{35.40} & \redbold{52.14} & \redbold{63.18} & \redbold{35.77} & \redbold{53.73} & \redbold{64.85} & \redbold{38.17} & \redbold{49.72} & \redbold{59.28} & \redbold{33.51} \\
\hline
\multirow{9}{*}{Scene}
& \cellcolor{lightpurple}TMBSD & 42.18 & 37.34 & 23.44 & 40.05 & 33.40 & 20.81 & 36.41 & 30.41 & 17.91 & 33.41 & 27.66 & 15.58 & 26.38 & 20.44 & 10.06 \\
& \cellcolor{lightpurple}TIMC & 42.37 & 41.20 & 25.20 & 38.06 & 34.51 & 19.80 & 32.79 & 27.80 & 14.53 & 28.10 & 23.13 & 11.34 & 23.61 & 19.79 & 08.57 \\
& \cellcolor{lightpurple}TCIMC & 29.93 & 27.71 & 13.72 & 33.30 & 30.05 & 15.71 & 35.65 & 32.16 & 17.23 & 36.32 & 31.77 & 17.82 & 33.19 & 27.51 & 14.71 \\
& \cellcolor{lightpurple}LATER & 64.09 & 61.83 & 49.99 & 60.86 & 57.04 & 45.07 & 55.37 & 50.54 & 37.94 & 48.00 & 44.18 & 30.34 & 37.83 & 34.78 & 20.53 \\
& \cellcolor{lightpurple}JPLTD & 81.73 & 84.12 & 76.71 & 79.44 & 81.29 & 72.77 & 78.00 & 79.82 & 70.46 & 76.54 & 78.49 & 67.91 & 75.54 & 72.75 & 68.56 \\
& \cellcolor{lightpurple}IMVCCBG & 30.21 & 26.39 & 14.26 & 30.39 & 26.51 & 14.33 & 30.24 & 26.46 & 14.29 & 30.20 & 26.44 & 14.27 & 30.21 & 26.43 & 14.28 \\
&  \cellcolor{lightpurple}ETLSRR & \bluebold{84.59} & \bluebold{83.59} & \bluebold{78.64} & \bluebold{84.24} & \bluebold{82.78} & \bluebold{77.69} & \bluebold{83.73} & \bluebold{81.78} & \bluebold{76.67} & \bluebold{83.54} & \bluebold{81.16} & \bluebold{76.01} & \bluebold{81.04} & \bluebold{76.67} & \bluebold{71.28} \\
& \cellcolor{lightpurple}Proposed & \redbold{86.15} & \redbold{87.24} & \redbold{82.64} & \redbold{85.41} & \redbold{85.89} & \redbold{81.03} & \redbold{85.05} & \redbold{85.28} & \redbold{80.30} & \redbold{84.73} & \redbold{84.80} & \redbold{79.64} & \redbold{82.22} & \redbold{81.50} & \redbold{75.97} \\
\hline
\multirow{9}{*}{Mfeat}
& \cellcolor{lightpurple}TMBSD & 95.97 & 91.87 & 91.70 & 94.16 & 88.02 & 87.90 & 91.59 & 83.52 & 82.91 & 89.03 & 80.73 & 79.05 & 85.92 & 77.31 & 74.89 \\
& \cellcolor{lightpurple}TIMC & 97.15 & 93.77 & 93.77 & 97.44 & 94.39 & 94.40 & 97.87 & 95.25 & 95.34 & 98.48 & 96.46 & 96.64 & 99.02 & 97.60 & 97.84 \\
& \cellcolor{lightpurple}TCIMC & 53.60 & 52.97 & 30.68 & 62.71 & 62.26 & 38.14 & 62.09 & 61.08 & 37.30 & 62.18 & 59.47 & 37.37 & 63.85 & 60.91 & 40.13 \\
& \cellcolor{lightpurple}LATER & 99.06 & 97.60 & 97.91 & 98.19 & 96.14 & 96.17 & 95.38 & 94.11 & 93.02 & 96.56 & 93.64 & 93.45 & 95.82 & 92.51 & 92.18\\
& \cellcolor{lightpurple}JPLTD & \redbold{99.90} & \redbold{99.74} & \redbold{99.78} & \redbold{99.87} & \redbold{99.66} & \redbold{99.72} & \redbold{99.81} & \redbold{99.50} & \redbold{99.58} & \redbold{99.77} & \redbold{99.41} & \redbold{99.50} & \redbold{99.75} & \redbold{99.34} & \redbold{99.44} \\
& \cellcolor{lightpurple}IMVCCBG & 73.36 & 67.81 & 60.30 & 73.42 & 67.82 & 60.33 & 73.85 & 68.04 & 60.63 & 73.73 & 67.97 & 60.53 & 73.86 & 68.04 & 60.61 \\
&  \cellcolor{lightpurple}ETLSRR & \bluebold{99.84} & \bluebold{99.55} & \bluebold{99.63} & \bluebold{99.83} & \bluebold{99.53} & \bluebold{99.62} & \bluebold{99.80} & \bluebold{99.46} & \bluebold{99.55} & \bluebold{99.75} & \bluebold{99.34} & \bluebold{99.46} & \bluebold{99.72} & 99.27 & \bluebold{99.39}  \\
& \cellcolor{lightpurple}Proposed & 99.47 & 98.67 & 98.83 & 99.31 & 98.26 & 98.46 & 99.40 & 98.43 & 98.66 & 99.66 & 99.12 & 99.24 & \bluebold{99.72} & \bluebold{99.28} & \bluebold{99.39} \\
\hline
\multirow{9}{*}{Caltech 7}
& \cellcolor{lightpurple}TMBSD & 38.58 & 39.61 & 28.10 & 37.85 & 34.68 & 26.32 & 37.60 & 30.31 & 22.91 & 36.99 & 26.93 & 20.99 & 36.85 & 27.53 & 21.50 \\
& \cellcolor{lightpurple}TIMC & \bluebold{62.97} & 55.56 & 50.61 & \bluebold{63.29} & 54.10 & 49.99 & \bluebold{62.41} & 53.49 & 49.66 & \bluebold{60.16} & 51.44 & 47.04 & \redbold{58.67} & 51.46 & \bluebold{46.72} \\
& \cellcolor{lightpurple}TCIMC & 44.25 & 39.54 & 29.40 & 44.40 & 39.74 & 29.67 & 46.20 & 39.04 & 30.96 & 43.83 & 38.77 & 28.83 & 45.20 & 37.54 & 28.86 \\
& \cellcolor{lightpurple}LATER & 60.26 & 58.20 & 48.44 & 58.89 & 58.10 & 48.33 & 58.05 & 56.14 & 47.21 & 57.43 & 55.02 & 46.74 & 55.92 & 53.43 & 44.99 \\
& \cellcolor{lightpurple}JPLTD & 60.73 & \bluebold{62.21} & \bluebold{52.15} & 60.89 & \bluebold{62.34} & \bluebold{52.15} & 60.74 & \bluebold{61.77} & \bluebold{51.77} & 59.59 & \bluebold{59.89} & \bluebold{49.74} & \bluebold{57.35} & \redbold{58.06} & \redbold{47.39} \\
& \cellcolor{lightpurple}IMVCCBG & 41.31 & 50.84 & 30.47 & 41.27 & 50.61 & 30.19 & 41.27 & 50.52 & 30.16 & 41.44 & 50.49 & 30.16 & 41.54 & 50.28 & 30.10 \\
&  \cellcolor{lightpurple}ETLSRR & 60.10 & 57.60 & 47.70 & 57.88 & 56.37 & 46.65 & 56.76 & 55.63 & 45.39 & 54.69 & 55.02 & 44.80 & \bluebold{54.56} & \bluebold{54.61} & 43.94 \\
& \cellcolor{lightpurple}Proposed & \redbold{70.24} & \redbold{67.42} & \redbold{60.17} & \redbold{68.44} & \redbold{66.37} & \redbold{58.56} & \redbold{65.75} & \redbold{65.21} & \redbold{56.60} & \redbold{62.67} & \redbold{62.81} & \redbold{53.74} & {55.02} & {52.02} & {43.39}  \\
\hline
\end{tabular}
\end{adjustbox}
\label{tab2}
\end{table*}

\subsection{Experimental results}

\begin{table*}[t]
\centering
\caption{ORL dataset: ablation study}
\begin{tabular}{ccccccccccccccccc}
\toprule
 \multirow{2}{*}{\textbf{Methods}} & \multicolumn{3}{c}{${p} = {0.1}$} & \multicolumn{3}{c}{${p} = {0.3}$} & \multicolumn{3}{c}{${p} = {0.5}$} & \multicolumn{3}{c}{${p} = {0.7}$} & \multicolumn{3}{c}{${p} = {0.9}$} \\
\cmidrule(lr){2-4} \cmidrule(lr){5-7} \cmidrule(lr){8-10} \cmidrule(lr){11-13} \cmidrule(lr){14-16}
&  \textbf{ACC} & \textbf{NMI} & \textbf{ARI} & \textbf{ACC} & \textbf{NMI} & \textbf{ARI} & \textbf{ACC} & \textbf{NMI} & \textbf{ARI} & \textbf{ACC} & \textbf{NMI} & \textbf{ARI} & \textbf{ACC} & \textbf{NMI} & \textbf{ARI} \\
\midrule
$\mathsf{L}_1$ & 83.74 & 92.72 & 79.12 & 88.05 & 94.94 & 85.07 & 79.05 & 89.16 & 71.99 & 78.04 & 87.74 & 69.31 & 75.16 & 85.42 & 64.21 \\
$\mathsf{L}_2$ & 84.08 & 93.01 & 79.78 & 88.38 & 95.17 & 85.72 & 78.84 & 89.15 & 71.88 & 78.06 & 87.90 & 69.57 & 75.01 & 85.37 & 64.11 \\
$\mathsf{L}_3$ & 78.81 & 89.25 & 71.32 & 77.04 & 88.06 & 68.96 & 74.82 & 85.33 & 63.75 & 66.21 & 78.34 & 49.06 & 60.27 & 74.35 & 42.42 \\
$\mathsf{L}_{1,2}$ & 91.90 & 97.15 & 90.77 & 92.25 & 97.15 & 91.01 & 92.88 & 97.19 & 91.46 & 86.88 & 94.28 & 83.84 & 84.13 & 92.79 & 80.07  \\
$\mathsf{L}_{1,3}$ & 87.86 & 94.64 & 84.24 & 83.61 & 92.18 & 78.28 & 81.93 & 90.77 & 75.61 & 82.01 & 90.69 & 75.86 & 81.60 & 89.57 & 73.10 \\
$\mathsf{L}_{2,3}$ & 87.88 & 94.60 & 84.22 & 83.93 & 92.30 & 78.60 & 81.88 & 90.64 & 75.33 & 82.89 & 90.60 & 75.63 & 81.96 & 89.76 & 73.58 \\
$\mathsf{L}_{1,2,3}$ & 96.68 & 99.02 & 96.59 & 95.89 & 98.62 & 95.51 & 95.35 & 98.36 & 94.78 & 94.72 & 98.06 & 93.95 & 84.43 & 91.31 & 79.20 \\
 Proposed & 98.38 & 99.53 & 98.35 & 97.87 & 99.18 & 97.47 & 96.76 & 98.74 & 96.13 & 95.75 & 98.21 & 94.72 & 94.84 & 97.91 & 93.72 \\
\bottomrule
\end{tabular}
\label{tab3}
\end{table*}

\begin{figure*}[!t]
    \centering
    \begin{minipage}[b]{0.33\textwidth}
        \includegraphics[width=\textwidth]{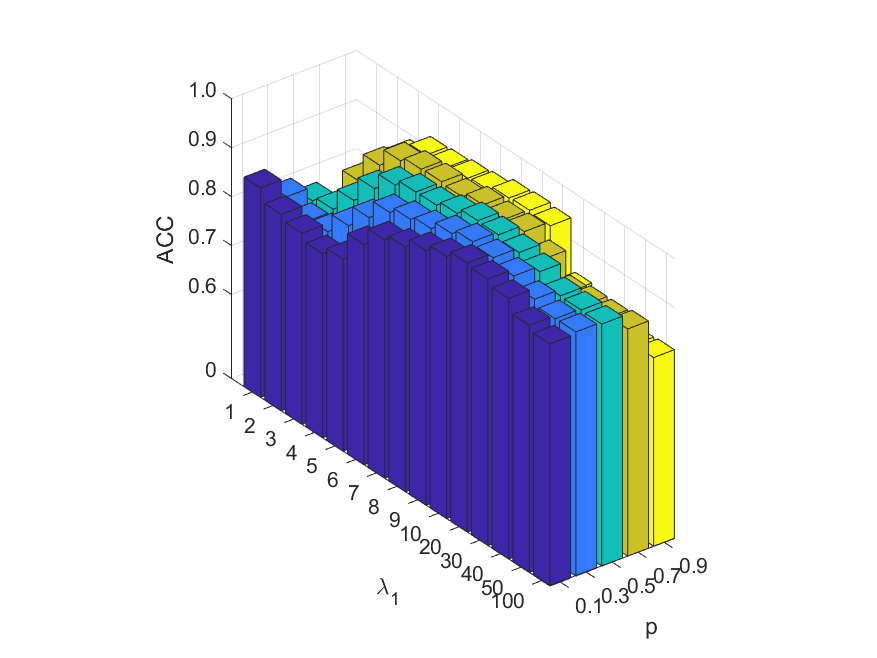}
    \end{minipage}
    \hspace{-1em}
    \begin{minipage}[b]{0.33\textwidth}
        \includegraphics[width=\textwidth]{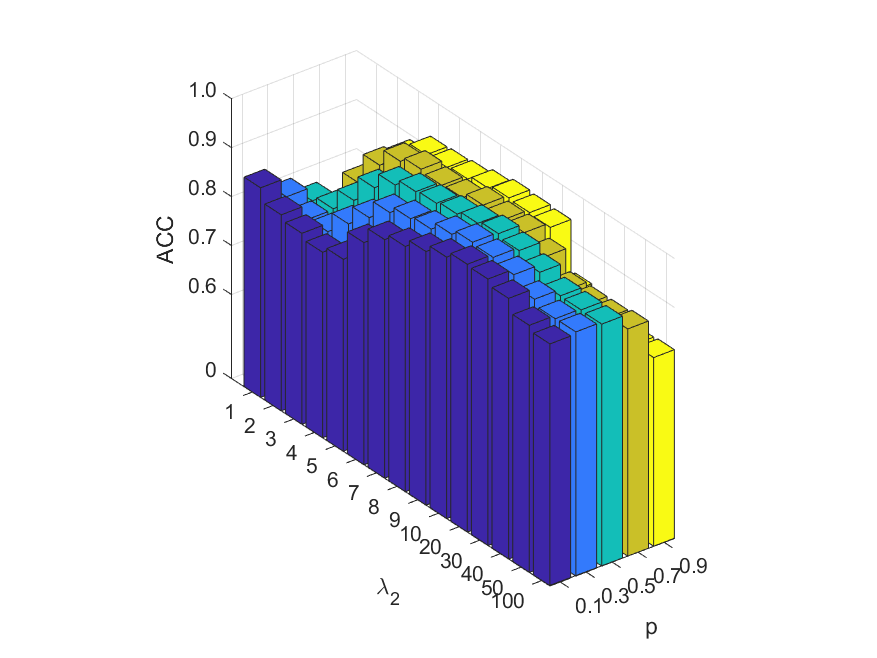}
    \end{minipage}
    \hspace{-1em}
    \begin{minipage}[b]{0.33\textwidth}
        \includegraphics[width=\textwidth]{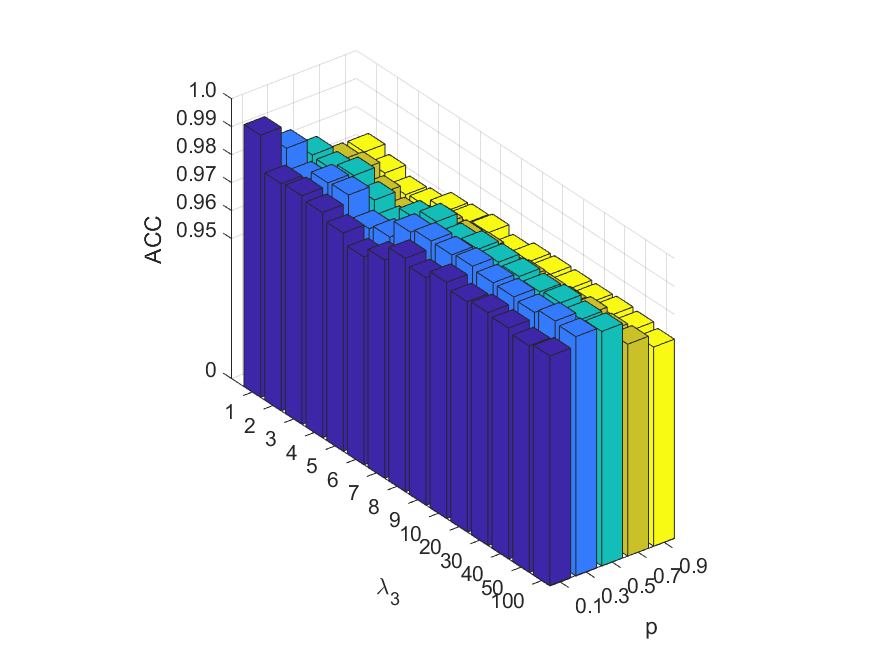}
    \end{minipage}
    \vspace{-0.1em}

    \begin{minipage}[b]{0.33\textwidth}
        \includegraphics[width=\textwidth]{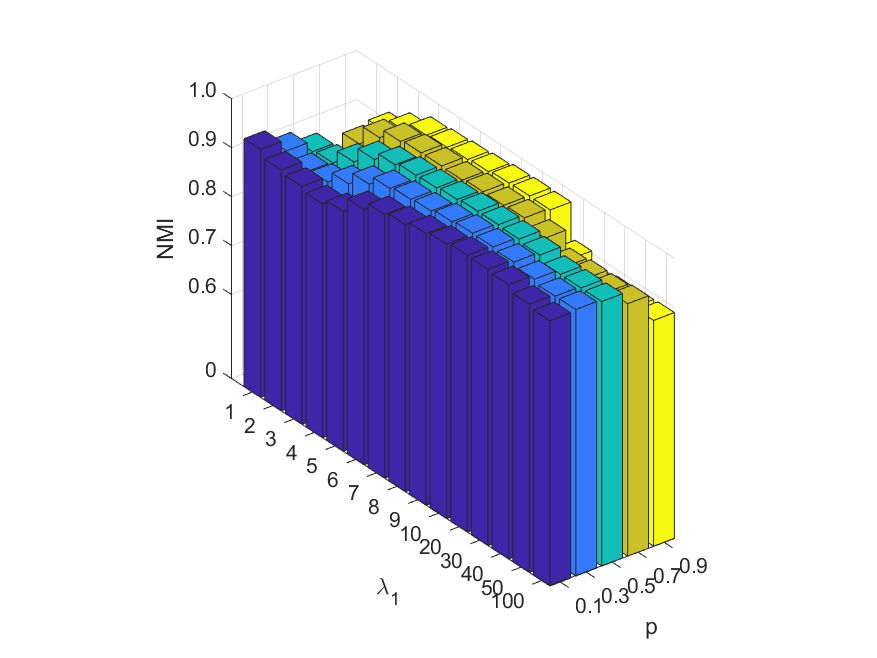}
    \end{minipage}
    \hspace{-1em}
    \begin{minipage}[b]{0.33\textwidth}
        \includegraphics[width=\textwidth]{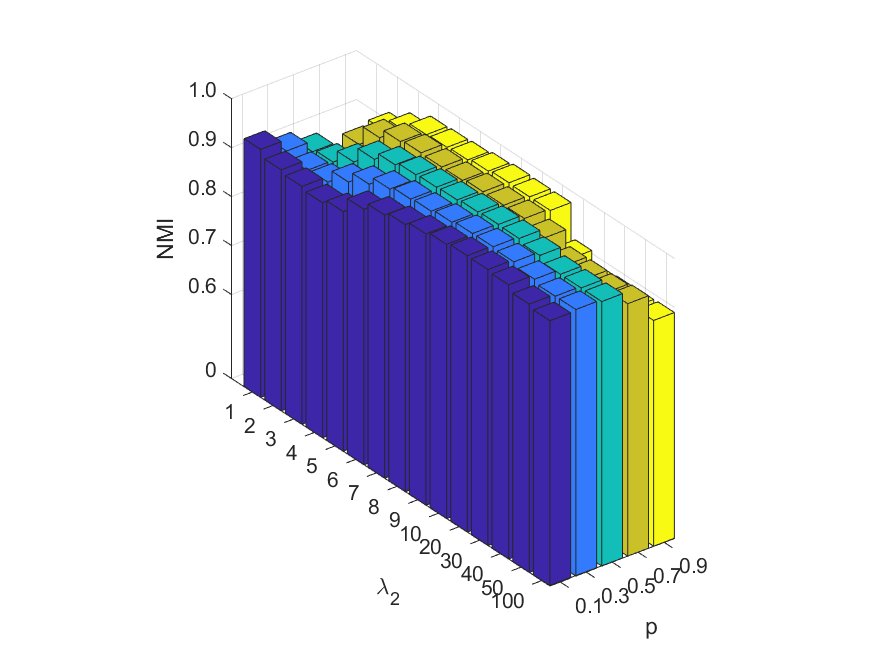}
    \end{minipage}
    \hspace{-1em}
    \begin{minipage}[b]{0.33\textwidth}
        \includegraphics[width=\textwidth]{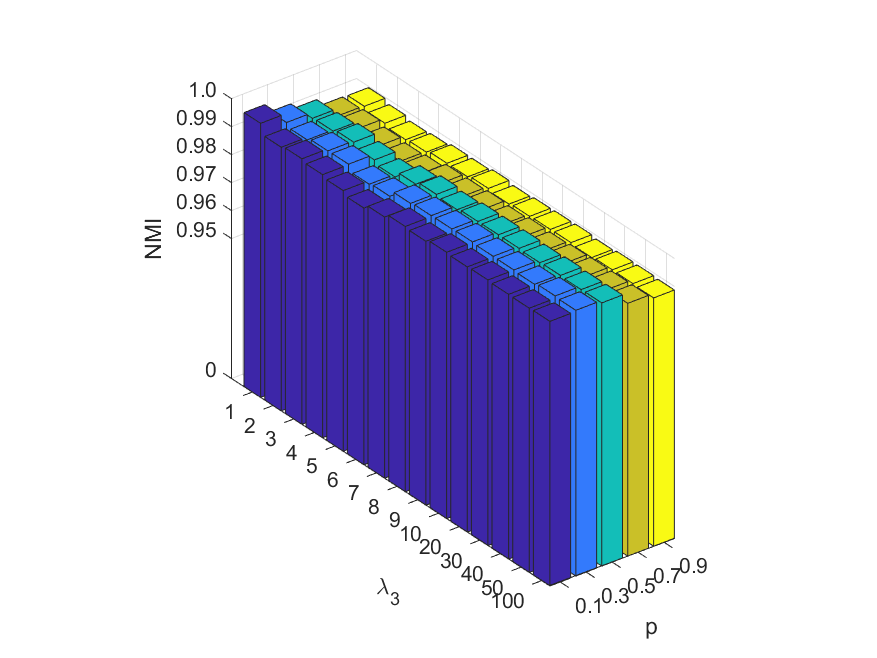}
    \end{minipage}
    \vspace{-0.1em}

    \begin{minipage}[b]{0.33\textwidth}
        \includegraphics[width=\textwidth]{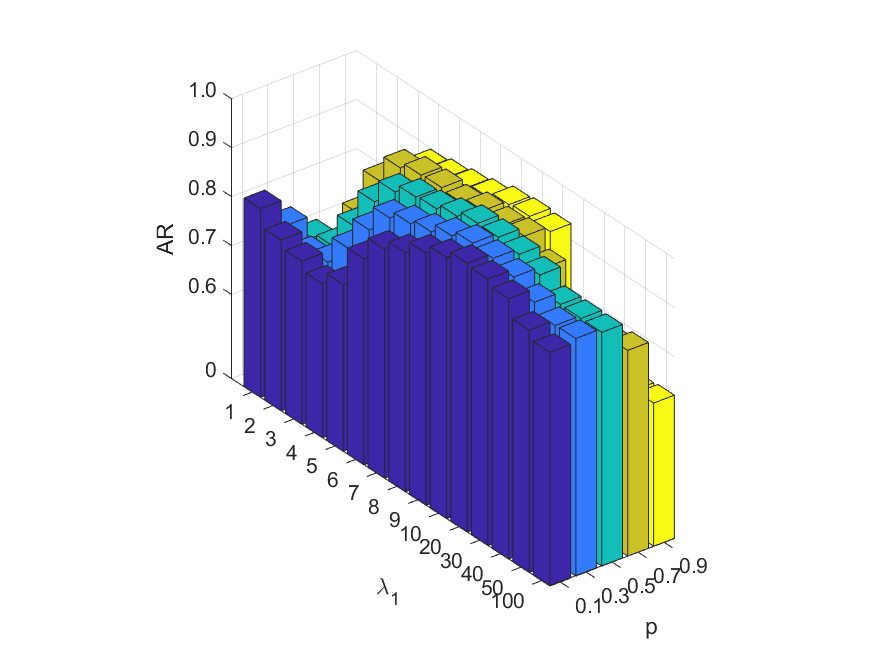}
    \end{minipage}
    \hspace{-1em}
    \begin{minipage}[b]{0.33\textwidth}
        \includegraphics[width=\textwidth]{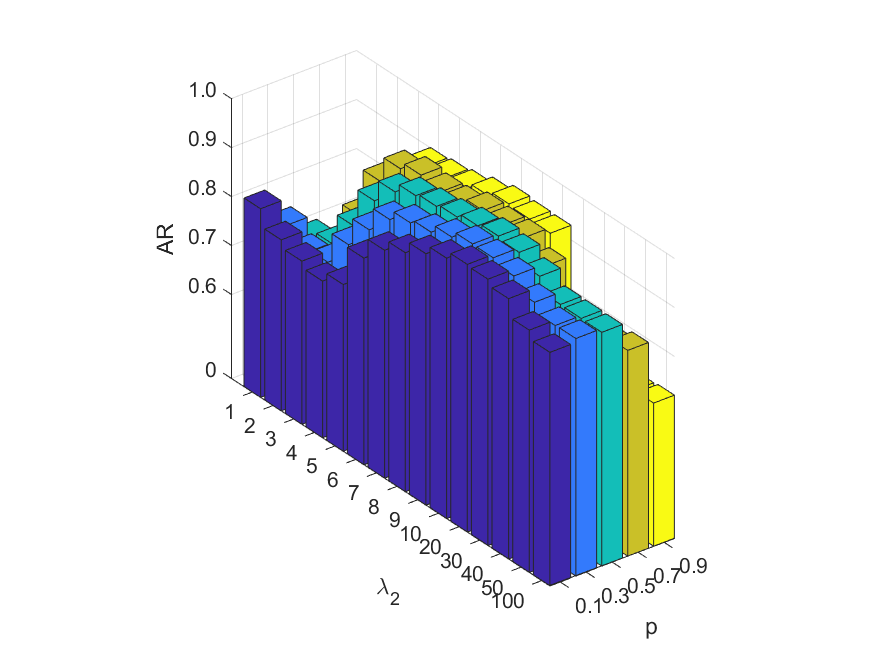}
    \end{minipage}
    \hspace{-1em}
    \begin{minipage}[b]{0.33\textwidth}
        \includegraphics[width=\textwidth]{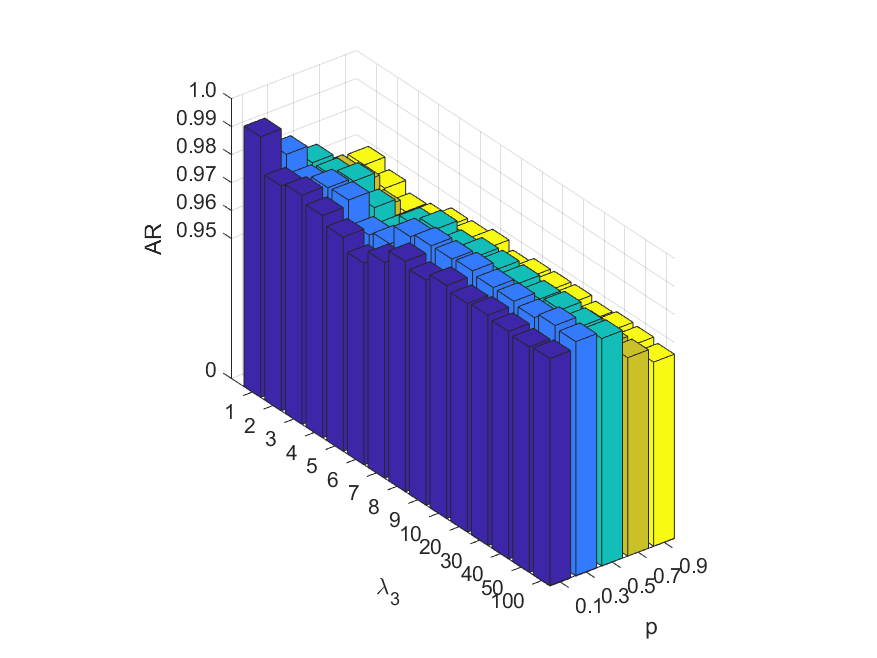}
    \end{minipage}

    \caption{ORL dataset: performance of the proposed JTIV-LRR method with respect to $\lambda_1$, ${\lambda_2}$, and ${\lambda_3}$ under various missing rates $p$.}
    \label{fig6}
\end{figure*}

\noindent \textbf{Visualization --} Figures \ref{fig4} and \ref{fig5} illustrate the recovered similarity graphs and their corresponding t-SNE visualizations for ETLSRR (top row), JPLTD (middle row) and the proposed JTIV-LRR method (bottom row) under varying missing rates $p$. In both figures, JTIV-LRR consistently outperforms ETLSRR and JPLTD in preserving structural information, retaining clear block-diagonal patterns (Fig. \ref{fig4}) and distinct class distributions (Fig. \ref{fig5}), even under extreme missing rates $p = 0.9$. ETLSRR exhibits significant degradations in both graph structures and class separability as the missing rate increases $p \geq 0.5$, with sparse and disorganized patterns in Fig. \ref{fig4} and overlapping clusters in Fig. \ref{fig5}. JPLTD demonstrates better resilience, partially preserving structural information at moderate missing rates, but struggles to maintain clarity at higher missing rates $p \geq 0.7$. Our method excels in maintaining distinct inter-class separation and cohesive intra-class compactness across all missing rates. This is evident in Fig. \ref{fig5}, where class clusters generated by our method remain well-separated and compact, compared to the overlapping and diffuse clusters produced by ETLSRR and JPLTD at higher missing rates.

\noindent \textbf{Clustering results -- }
Table \ref{tab2} reports the clustering performance of the compared methods, namely, TMBSD, TIMC, TCIMC, LATER, JPLTD, ETLSRR and the proposed JTIV-LRR method, across multiple datasets under various missing rates $p$. The evaluation metrics include ACC, NMI, and ARI, with red bold highlighting the best performance and blue bold indicating the second-best. The datasets range from simpler cases (e.g., 3 sources and 100 leaves) to more complex ones (e.g., ProteinFold and Scene), testing the robustness and adaptability of the methods in handling incomplete multiview data.

These results demonstrate that the proposed JTIV-LRR method consistently achieves the best performance across almost all datasets and missing rates, particularly excelling under high missing rates $p = 0.7, 0.9$. While JPLTD and ETLSRR exhibit competitive performance under lower missing rates $p=0.1, 0.3, 0.5$, their effectiveness diminishes significantly as the missing rate increases. In contrast, our method maintains robust clustering quality, demonstrating superior inter-class separation and intra-class compactness, even on challenging datasets like ProteinFold and Scene. These results validate the robustness and generalizability of our approach in incomplete multiview clustering tasks.\\

\begin{figure*}[t]
    \centering

    \begin{tikzpicture}
        \node[anchor=south west, inner sep=0] (title1) at (-6.5, 0) {$p=0.1$};
        \node[anchor=south west, inner sep=0] (title2) at (-3.0, 0) {$p=0.3$};
        \node[anchor=south west, inner sep=0] (title3) at (0.5, 0) {$p=0.5$};
        \node[anchor=south west, inner sep=0] (title4) at (3.75, 0) {$p=0.7$};
        \node[anchor=south west, inner sep=0] (title5) at (7.25, 0) {$p=0.9$};
    \end{tikzpicture}

    \begin{tikzpicture}[baseline]

        \node[inner sep=0] (col1) at (0.0, 0) {
            \begin{minipage}[b]{0.19\textwidth}
                \includegraphics[width=\textwidth]{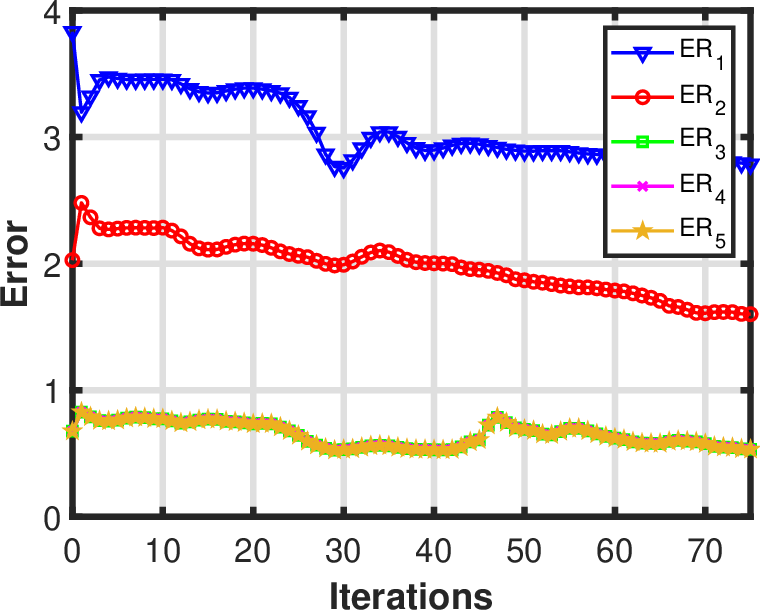} \\[-0.5em]
                \includegraphics[width=\textwidth]{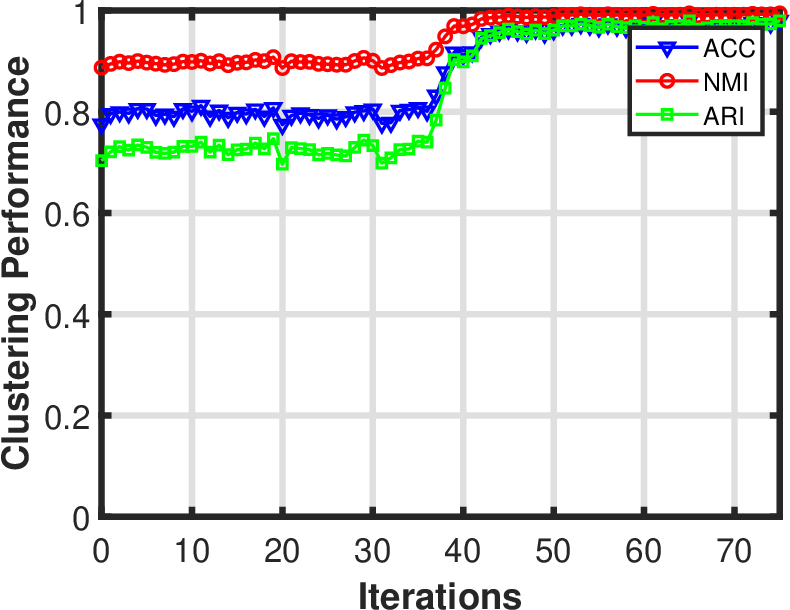}
            \end{minipage}
        };

        \node[inner sep=0] (col2) at (0.19\textwidth, 0) {
            \begin{minipage}[b]{0.19\textwidth}
                \includegraphics[width=\textwidth]{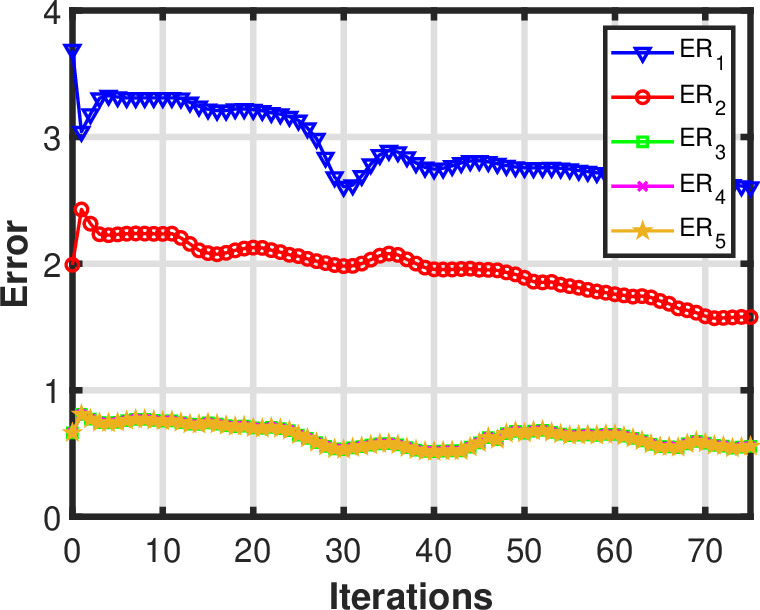} \\[-0.5em]
                \includegraphics[width=\textwidth]{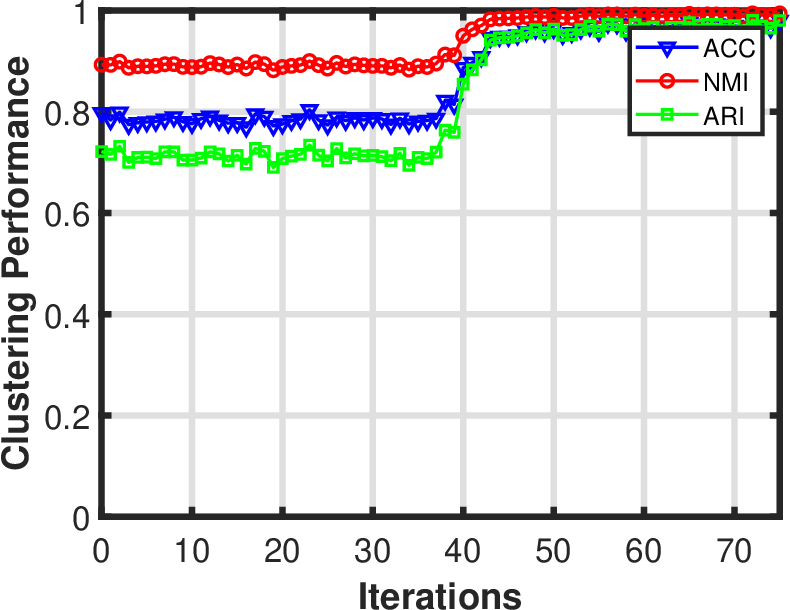}
            \end{minipage}
        };

        \node[inner sep=0] (col3) at (0.38\textwidth, 0) {
            \begin{minipage}[b]{0.19\textwidth}
                \includegraphics[width=\textwidth]{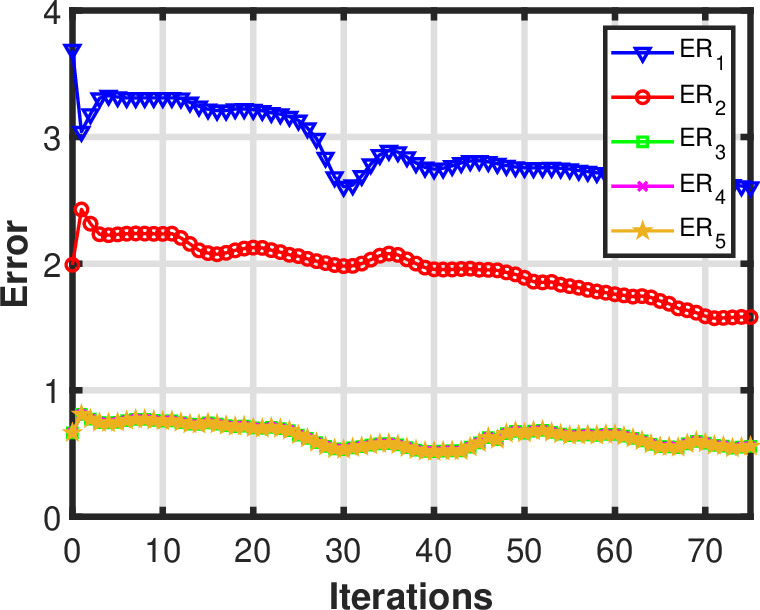} \\[-0.5em]
                \includegraphics[width=\textwidth]{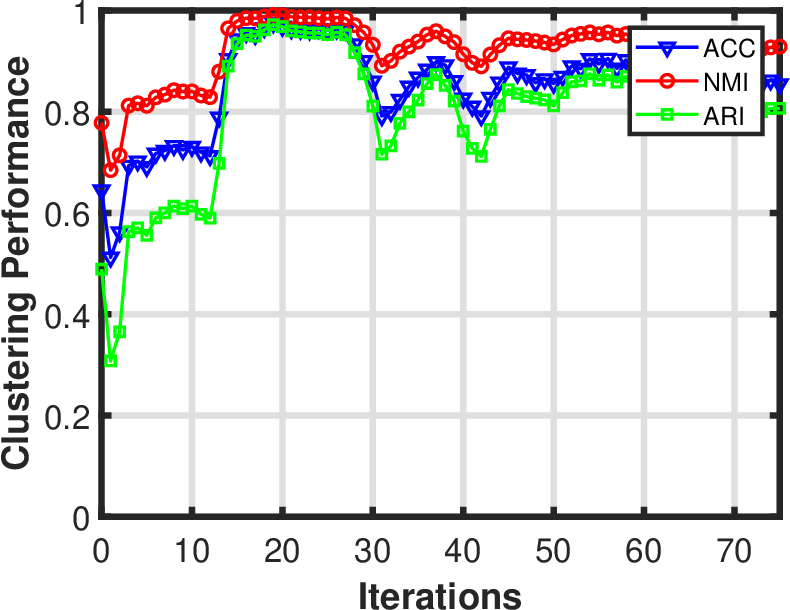}
            \end{minipage}
        };

        \node[inner sep=0] (col4) at (0.57\textwidth, 0) {
            \begin{minipage}[b]{0.19\textwidth}
                \includegraphics[width=\textwidth]{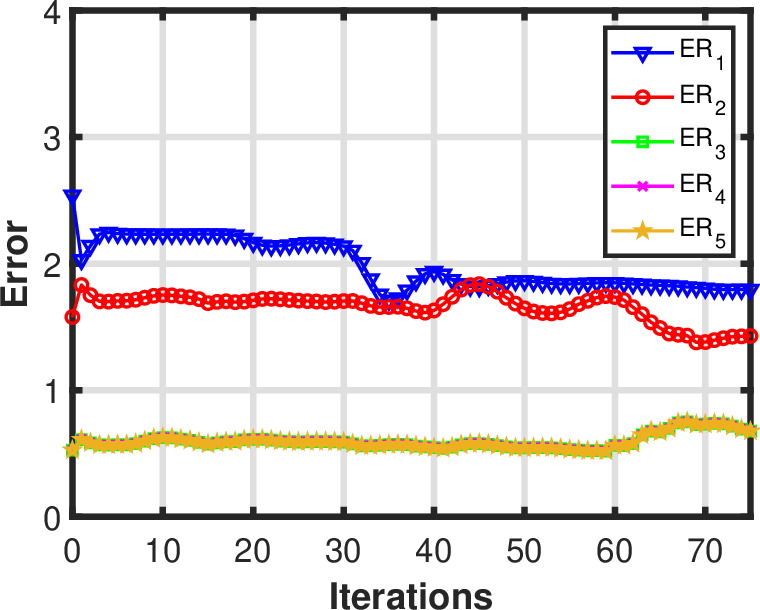} \\[-0.5em]
                \includegraphics[width=\textwidth]{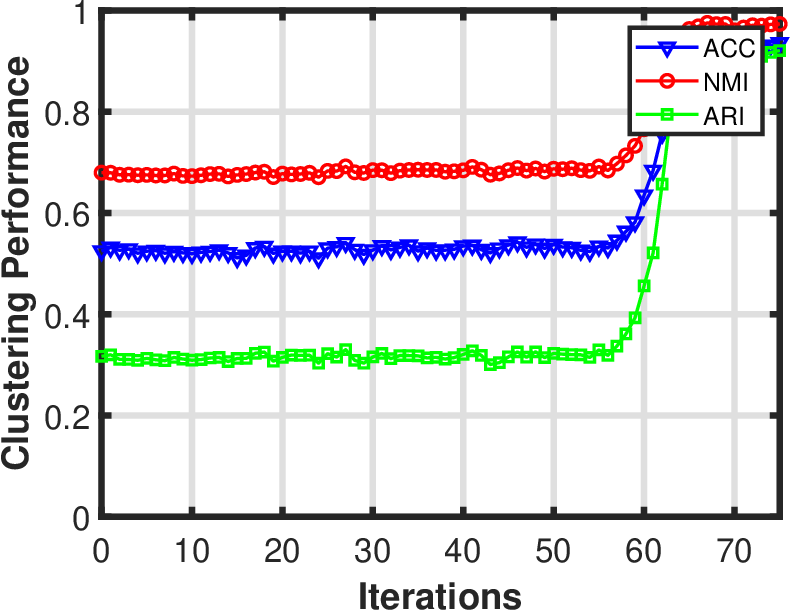}
            \end{minipage}
        };

        \node[inner sep=0] (col5) at (0.76\textwidth, 0) {
            \begin{minipage}[b]{0.19\textwidth}
                \includegraphics[width=\textwidth]{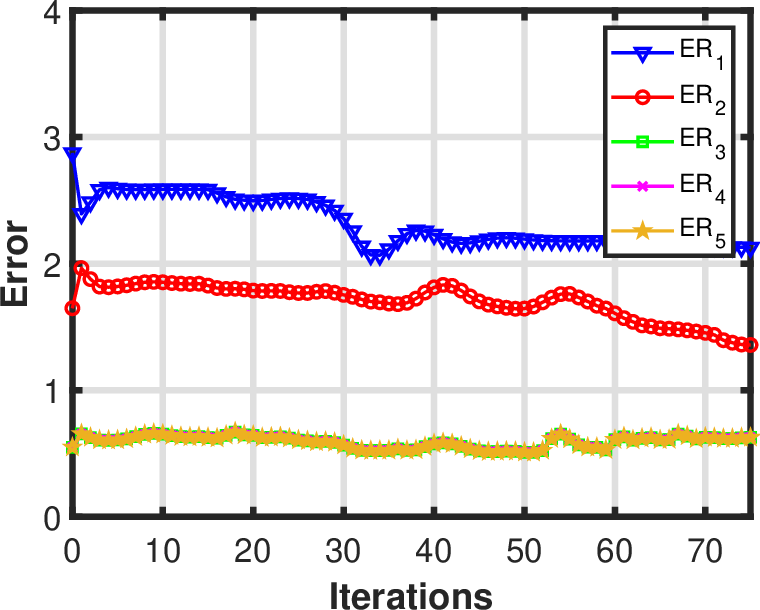} \\[-0.5em]
                \includegraphics[width=\textwidth]{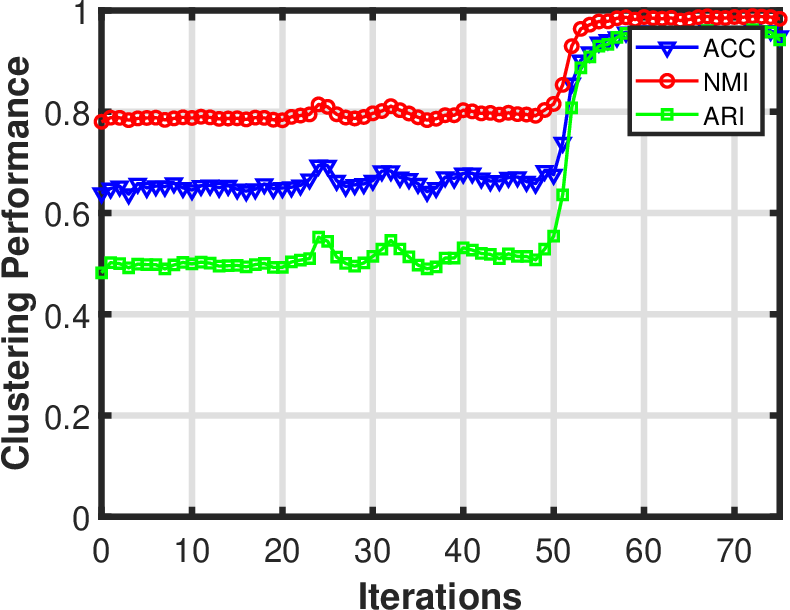}
            \end{minipage}
        };

    \end{tikzpicture}

    \caption{ORL dataset: convergence the proposed JTIV-LRR method under various missing rates $p$.}
    \label{fig7}
\end{figure*}

\noindent \textbf{Ablation study -- }
This ablation study evaluates the impact of the different terms composing the objective function on the performance of the proposed JTIV-LRR method under varying missing rates $p = \{0.1, 0.3, 0.5, 0.7, 0.9\}$ on the ORL dataset. Specifically, in what follows $\mathsf{L}_1$ denotes a depreciated counterpart of the proposed approach by applying only a low-tubal-rank penalization to $\mathcal{L}$; $\mathsf{L}_2$ applies this penalization to $\mathcal{L}_{[2]}$, and $\mathsf{L}_3$ applies it to $\mathcal{L}_{[3]}$. When combining pair of low-rank penalizations to $\mathcal{L}_{[m]}$ and $\mathcal{L}_{[j]}$, the corresponding methods are denoted as $\mathsf{L}_{m,j}$ with $(m,j) \in \{(1,2),(1,3),(2,3)\}$. Finally and $\mathsf{L}_{1,2,3}$, include the respective penalizations on the three multiple mode permutations of $\mathcal{L}$, while removing the sparse promoting regularization. One recall that the proposed JTIV-LRR method incorporates the three low-tubal-rank penalizations as $\mathsf{L}_{1,2,3}$ but also  the sparse promoting regularization.

The results obtained on  the ORL dataset and reported in Table \ref{tab3} demonstrate that incorporating sparse noise constraints significantly enhances performance, as evidenced by the consistent superiority of our method over its variant without sparse noise constraints ($\mathsf{L}_{1,2,3}$). Additionally, combining low-tubal-rank constraints across multiple matrix permutations ($\mathsf{L}_{1,2,3}$) consistently outperforms using individual constraints ($\mathsf{L}_1$, $\mathsf{L}_2,$ or $\mathsf{L}_3$), highlighting the effectiveness of the proposed inter-view low-rank constraints. The proposed method achieves the best results across all metrics (ACC, NMI, ARI) and missing rates, demonstrating its robustness and effectiveness in capturing multi-view low-rank structures while mitigating sparse noise.\\

\noindent  \textbf{Parameters analysis --}
This set of experiments evaluates the performance of the model under varying parameters ${\lambda_1}, {\lambda_2}, {\lambda_3}$ and missing rates $p$ using the ORL dataset. The parameters are tested across the range $\{1,2,3,4,5,6,7,8,9,10,20,30,40,50,100\}$. Model performance is assessed using ACC, NMI, and ARI, aiming to analyze the sensitivity of the model to parameter tuning and its robustness against missing data.

As shown in Fig. \ref{fig6}, the model achieves stable and high performance, $\textrm{ACC} > 0.95$, under low missing rates $p<0.5$, while NMI and ARI exhibit a notable decline at higher missing rates $p>0.6$. Among the parameters, ${\lambda_1}$ and ${\lambda_2}$ significantly affect performance, with optimal values observed in the mid-range, whereas ${\lambda_3}$ have relatively minor impact but require careful adjustment to avoid extreme values. Overall, the findings highlight the importance of parameter tuning in maintaining robust performance under high missing rate scenarios.\\

\noindent   \textbf{Convergence analysis --}
The algorithmic convergence of the proposed method is evaluated by monitoring five residuals intrinsic to ADMM and defined as
\begin{align}
    \mbox{ER}_1 &\defeq \max_v \left\{ \left\| \mathbf{X}^{(v)} - \mathbf{X}^{(v)} \mathbf{W}^{(v)^\mathsf{T}} \tilde{\mathbf{G}}^{(v)} \mathbf{W}^{(v)} \right\|_\infty \right\} \nonumber\\
    \mbox{ER}_2 &\defeq \left\| \tilde{\mathcal{G}} - \tilde{\mathcal{L}} - \tilde{\mathcal{S}} \right\|_{\infty},  \quad \mbox{ER}_3 \defeq \left\| \tilde{\mathcal{L}} - \mathcal{Z}_1 \right\|_\infty\nonumber\\
    \mbox{ER}_4 &\defeq \left\| \tilde{\mathcal{L}}_{[2]} - \mathcal{Z}_2 \right\|_\infty, \quad \ \mbox{ER}_5 \defeq \left\| \tilde{\mathcal{L}}_{[3]} - \mathcal{Z}_3 \right\|_\infty. \nonumber
\end{align}
Figure \ref{fig7} shows the corresponding residuals as well as the clustering performance along the algorithm iterations on the ORL dataset under various missing rates. One can observe that as the iteration progresses, the residuals consistently decreases across all missing rates $p$ and eventually converges, indicating the effectiveness of the optimization process. However, as the missing rate $p$ increases ($p=0.7$ and $p=0.9$), the error remains higher, and the clustering performance metrics (ACC, NMI, ARI) degrade significantly, showing the sensitivity of the algorithm to high missing rates. In contrast, at lower missing rates ($p=0.1$, and $p=0.3$), the algorithm achieves lower errors and higher clustering performance, demonstrating its robustness in scenarios with fewer missing values.

\section{Conclusion}
This paper proposes JTIV-LRR, a novel incomplete multi-view clustering algorithm that unifies incomplete similarity graph learning and complete tensor representation recovery within a single framework. By incorporating low-tubal-rank constraints across multiple views, the method effectively captures both consistency and complementarity among views, ensuring more robust and accurate clustering. Extensive experiments on synthetic and seven real-world datasets validate the effectiveness and generalizability of the proposed approach, consistently surpassing state-of-the-art methods such as ETLSRR and JPLTD in clustering performance.

\bibliographystyle{IEEEbib}
\bibliography{strings,refs}

\vfill

\end{document}